%% file: main.tex
\documentclass{article}

\usepackage{microtype}
\usepackage{graphicx}
\usepackage{subfigure}
\usepackage{booktabs} % for professional tables

\usepackage{hyperref}

\usepackage[accepted]{icml2024}

\usepackage{amsmath}
\usepackage{amssymb}
\usepackage{mathtools}
\usepackage{amsthm}

\usepackage[capitalize,noabbrev]{cleveref}

\theoremstyle{plain}

\theoremstyle{definition}

\theoremstyle{remark}

\usepackage[textsize=tiny]{todonotes}

\input{preamble.tex}

\icmltitlerunning{Superpoint Gaussian Splatting for Real-Time High-Fidelity Dynamic Scene Reconstruction}
\begin{document}
\twocolumn[
\icmltitle{Superpoint Gaussian Splatting for Real-Time High-Fidelity Dynamic Scene Reconstruction}




\begin{icmlauthorlist}
\icmlauthor{Diwen Wan}{pku}
\icmlauthor{Ruijie Lu}{pku}
\icmlauthor{Gang Zeng}{pku}
\end{icmlauthorlist}

\icmlaffiliation{pku}{National Key Laboratory of General Artificial Intelligence, School of IST, Peking University, China}

% \icmlcorrespondingauthor{Diwen Wan}{wan@stu.pku.edu.cn}
\icmlcorrespondingauthor{Gang Zeng}{zeng@pku.edu.cn}

\icmlkeywords{3D Reconstruction, Novel View Synthesis, Dynamic Scene, Gaussian Splatting}

\vskip 0.3in
]



\printAffiliationsAndNotice{}  % leave blank if no need to mention equal contribution

\input{sec/0_abstract.tex}
\input{sec/1_intro.tex}
\input{sec/2_related.tex}
\input{sec/3_method.tex}
\input{sec/4_experiment.tex}
\input{sec/5_applications.tex}
\input{sec/6_conclusions.tex}
\input{sec/7_other.tex}


\bibliography{bib/cj_short_name,bib/3D_NeRF,bib/3D_other,bib/3D_reconstruction,bib/superpixel_point,bib/networks}
\bibliographystyle{icml2024}

%%%%%%%%%%%%%%%%%%%%%%%%%%%%%%%%%%%%%%%%%%%%%%%%%%%%%%%%%%%%%%%%%%%%%%%%%%%%%%%
%%%%%%%%%%%%%%%%%%%%%%%%%%%%%%%%%%%%%%%%%%%%%%%%%%%%%%%%%%%%%%%%%%%%%%%%%%%%%%%
% APPENDIX
%%%%%%%%%%%%%%%%%%%%%%%%%%%%%%%%%%%%%%%%%%%%%%%%%%%%%%%%%%%%%%%%%%%%%%%%%%%%%%%
%%%%%%%%%%%%%%%%%%%%%%%%%%%%%%%%%%%%%%%%%%%%%%%%%%%%%%%%%%%%%%%%%%%%%%%%%%%%%%%
\newpage
\appendix
% \onecolumn
\input{sec/X_suppl.tex}
%%%%%%%%%%%%%%%%%%%%%%%%%%%%%%%%%%%%%%%%%%%%%%%%%%%%%%%%%%%%%%%%%%%%%%%%%%%%%%%
%%%%%%%%%%%%%%%%%%%%%%%%%%%%%%%%%%%%%%%%%%%%%%%%%%%%%%%%%%%%%%%%%%%%%%%%%%%%%%%

\end{document}

%% file: preamble.tex
\usepackage{multicol}
\usepackage{multirow}
\usepackage{booktabs}
\usepackage{graphicx}
\usepackage{amssymb}
\usepackage{amsthm}
\usepackage{amsmath}
\usepackage{bm}
\usepackage{caption}
\usepackage{subcaption}

\usepackage[export]{adjustbox}
\newcommand{\inR}[1]{\in \mathbb{R}^{#1}}

\DeclareMathOperator*{\argmax}{\arg\max}

\newcommand{\gsR}{\mathbf{R}}
\newcommand{\gsP}{\bm{\mu}}

\newcommand{\spR}{\Delta\mathbf{R}}
\newcommand{\spT}{\Delta\bm{t}}
\newcommand{\spP}{\bm{p}}
\newcommand{\knn}{\mathcal{N}}

\newcommand{\spRg}{\mathrm{R}_{sp \to g}}
\newcommand{\gRsp}{\mathrm{R}_{g \to sp}}

\newcommand{\SP}{\mathbb{S}}

\newcommand{\se}{\mathfrak{s}\mathfrak{e}}

\newcommand{\Asso}{\mathbf{A}} % Assocication map
\newcommand{\loss}{\mathcal{L}}

\definecolor{c1}{HTML}{FFB2B2}
\definecolor{c2}{HTML}{FFD9B2}
\definecolor{c3}{HTML}{F2F2F2}
\newcommand{\colorTop}[2][1]{%
  \ifnum#1=1%
\colorbox{c1}{\textbf{#2}}%
\else \ifnum#1=2%
\colorbox{c2}{\underline{#2}}%
\else \ifnum#1=3%
\colorbox{c3}{#2}% 
\else %
#2%
\fi  \fi  \fi
}

\usepackage{xspace}
\makeatletter
\DeclareRobustCommand\onedot{\futurelet\@let@token\@onedot}
\def\@onedot{\ifx\@let@token.\else.\null\fi\xspace}

\def\ie{\emph{i.e}\onedot}

\makeatother

\usepackage{tabularx}
\usepackage{pifont}

%% file: sec/0_abstract.tex
%mainfile: main.tex
\begin{abstract}
    Rendering novel view images in dynamic scenes is a crucial yet challenging task. 
    Current methods mainly utilize NeRF-based methods to represent the static scene and an additional time-variant MLP to model scene deformations, resulting in relatively low rendering quality as well as slow inference speed. 
    To tackle these challenges, we propose a novel framework named Superpoint Gaussian Splatting (SP-GS). 
    Specifically, our framework first employs explicit 3D Gaussians to reconstruct the scene and then clusters Gaussians with similar properties (e.g., rotation, translation, and location) into superpoints. Empowered by these superpoints, our method manages to extend 3D Gaussian splatting to dynamic scenes with only a slight increase in computational expense. 
    Apart from achieving state-of-the-art visual quality and real-time rendering under high resolutions, the superpoint representation provides a stronger manipulation capability.
    Extensive experiments demonstrate the practicality and effectiveness of our approach on both synthetic and real-world datasets. Please see our project page at \url{https://dnvtmf.github.io/SP_GS.github.io}.
\end{abstract}

%% file: sec/1_intro.tex
\section{Introduction}
\label{sec:intro}

Synthesizing high-fidelity novel view images of a 3D scene is imperative for various industrial applications, ranging from gaming and filming to AR/VR. In recent years, Neural Radiance Fields(NeRF) ~\cite{NeRF} has demonstrated its remarkable ability on this task with photorealistic renderings. While lots of subsequent works focus on improving rendering quality~\cite{Mip-NeRF, Mip-NeRF360} or training and rendering speed~\cite{Instant-NGP, TensoRF, EfficientNeRF, Plenoxels} for static scenes, another line of work~\cite{D-NeRF, KPlanes, TiNeuVox} proposes to extend the setting to a dynamic scene. Though various attempts have been made to improve efficiency and dynamic rendering quality, the introduction of an additional time-variant MLP to model complex motions in dynamic scenes will inevitably cause a surge in computational cost during both the training and inference process.

More recently, 3D Gaussian Splatting(3D-GS)~\cite{3D-GS} manages to achieve real-time rendering with high visual quality by introducing a novel point-like representation, referred to as 3D Gaussians. However, it mainly deals with static scenes. Though methods such as leveraging a deformation network on each 3D Gaussian can extend 3D-GS to dynamic scenes, the rendering speed will be greatly affected, especially when a large number of 3D Gaussians is necessary to represent the scene.

Drawing inspiration from the well-established \textit{As-Rigid-As-Possible} regularization in 3D reconstruction and the superpoint/superpixel concept in point clouds/images over-segmentation, we propose a novel approach named Superpoint Gaussian Splatting (SP-GS) in this work for reconstructing and rendering dynamic scenes. 
The key insight lies in that each 3D Gaussian should not be a completely independent entity. Some neighboring 3D Gaussians probably possess similar translation and rotation transformations at all timesteps due to the properties of a rigid motion. We can cluster these similar 3D Gaussians together to form a superpoint so that it is no longer necessary to compute a deformation for every single 3D Gaussian, leading to a much faster rendering speed.

To be specific, after acquiring the initial 3D Gaussians of the canonical space through a warm-up training process, a learnable association matrix will be applied to the initial 3D Gaussians and group them into several superpoints. Subsequently, our framework will leverage a tiny MLP network for predicting the deformations of superpoints, which will later be utilized to compute the deformation of every single 3D Gaussian in the superpoint to enable novel view rendering for dynamic scenes. Apart from the rendering loss at each timestep, to take full advantage of the 
 \textit{As-Rigid-As-Possible} feature within one superpoint, we additionally utilize a property reconstruction loss on the properties of Gaussians, including positions, translations, and rotations. 

Thanks to the computational expense saved by using superpoints, our approach manages to achieve a comparable rendering speed with 3D-GS. Furthermore, the mixed representation of 3D Gaussians and superpoints possess a strong extensibility like adding a non-rigid motion prediction module for better dynamic scene reconstruction. Last but not least, SP-GS can facilitate various downstream applications like editing a reconstructed scene as superpoints should cluster similar 3D Gaussians together, providing more meaningful groups than pure 3D Gaussians. 
Our contributions can be summarized as follows:
\begin{itemize}
  \item We introduce Superpoint Gaussian Splatting (SP-GS), a novel approach for high-fidelity and real-time rendering in dynamic scenes that aggregates 3D Gaussians with similar deformations into superpoints. 
  \item Our method possesses a strong extensibility like adding a non-rigid prediction module or distillation from a larger model and can facilitate various downstream applications like scene editing.
  \item SP-GS achieves real-time rendering on dynamic scenes, up to 227 FPS at a resolution of $800\times 800$ for synthetic datasets and 117 FPS at a resolution of $536 \times 960$ in real datasets with superior or comparable performance than previous SOTA methods.
\end{itemize}
% See our project page for video results and source code at \url{https://dnvtmf.github.io/SP_GS.github.io}.

%% file: sec/2_related.tex
\section{Related Works}
\subsection{Static Neural Rendering}
In recent years, we have witnessed significant progress in the field of novel view synthesis empowered by Neural Radiance Fields. While vanilla NeRF~\cite{NeRF} manages to synthesize photorealistic images for any viewpoint using MLPs, numerous subsequent works focus on acceleration~\cite{Plenoxels, EfficientNeRF, SNeRF, Instant-NGP}, real-time rendering~\cite{MobileNeRF, PlenOctrees}, camera parameter optimization~\cite{NoPe-NeRF, BARF, FF-NeRF}, few-shot learning~\cite{FDNeRFF, FreeNeRF, pixelNeRF}, unbounded scenes~\cite{Mip-NeRF360, Omni-NeRF}, improving visual quality~\cite{Mip-NeRF, Zip-NeRF}, and so on.

More recently, a novel framework 3D Gaussian Splatting~\cite{3D-GS} has received widespread attention for its ability to synthesize high-fidelity images for complex scenes in real-time along with a fast training speed. The key insight lies in that it exploits a point-like representation, referred to as 3D Gaussians. However, these works are mainly restricted to the domain of static scenes.

\subsection{Dynamic Neural Rendering}

To extend neural rendering to dynamic scenes, current efforts primarily focus on deformation-based~\cite{D-NeRF, HyperNeRF, NR-NeRF} and flow-based methods ~\cite{DyNeRF, NSFF, NeRFlow, VideoNeRF}. However, these approaches share similar issues as NeRF, including slow training and rendering speed. To mitigate the efficiency problem, various acceleration techniques like voxel~\cite{TiNeuVox, DeVRF} or hash-encoding representation~\cite{TI-DNeRF}, and spatial decomposition~\cite{KPlanes, Tensor4D, HexPlane, D2NeRF} have emerged. Given the increased complexity brought by dynamic scene modeling, there still exists a gap in rendering quality, training time, and rendering speed compared to static scenes. 

Concurrent with our work, methods like Deformable 3D Gaussians (D-3D-GS)~\cite{Deformable3DGS}, 4D-GS~\cite{RealTime4DGS}, and Dynamic 3D Gaussians~\cite{Dynamic3DGS} leverage 3D-GS as the scene representation, expecting this novel point-like representation can facilitate dynamic scene modeling. D-3D-GS directly integrates a heavy deformation network into 3D-GS, while 4D-GS combines HexPlane~\cite{HexPlane} with 3D-GS to achieve real-time rendering and superior visual quality. Dynamic 3D Gaussians proposes a method that simultaneously addresses the tasks of dynamic scene novel-view synthesis and 6-DOF tracking of all dense scene elements. While our method also takes 3D-GS as the scene representation, unlike any of the aforementioned methods, our main motivation is to aggregate 3D Gaussians with similar deformations into a superpoint to significantly decrease the computational expense required.

\subsection{Superpixel/Superpoint}
There exists a long line of research works on superpixel/superpoint segmentation and we refer readers to the recent paper~\cite{J2023AnES} for a thorough survey. Here we focus on neural network-based methods. 

On one hand, methods including SFCN~\cite{SFCN}, AINet~\cite{AINet}, and LNS-Net~\cite{LNS-Net} adopt a neural network for generating superpixels. SFCN utilizes a fully convolutional network associated with an SLIC loss, while AINet introduces an implantation module and a boundary-perceiving SLIC loss for generating superpixels with more accurate boundaries. LNS-Net proposes an online learning framework, alleviating the demand for large-scale manual labels. On the other hand, existing methods for point cloud over-segmentation can be divided into two categories: optimization-based methods~\cite{Papon2013VoxelCC, Lin2018TowardBB, Guinard2017WEAKLYSS, Landrieu2016CutPF} and deep learning-based methods ~\cite{Landrieu2019PointCO, Hui_2023_ICCV, Hui2021SuperpointNF}.

Our approach can be treated as an over-segmentation of dynamic point clouds, which is an unexplored realm. Existing superpixel/superpoint methods cannot be directly applied to our task since it is challenging to maintain superpoint-segmentation consistency across the temporal domain. Moreover, prevalent methods either employ computationally intensive backbones or do not support parallelization, making the segmentation a heavy module, which will hinder the fast training and rendering speed of our approach.

%% file: sec/3_method.tex
\section{Methods}
\begin{figure*}
    \centering 
    \includegraphics[width=0.9\linewidth]{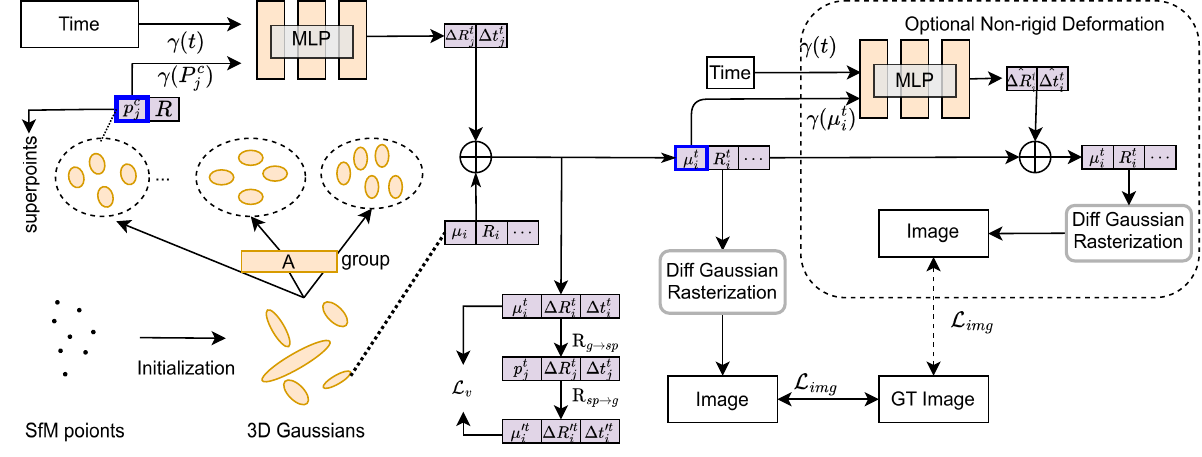} 
    \caption{\textbf{Overview of our pipeline.} We initialize the 3D Gaussians with point clouds reconstructed from SfM. Then we aggregate the 3D Gaussians into superpoints, and predict the deformation for every 3D Gaussian at a given timestep. The image is rendered using the differentiable Gaussian rasterization on the deformed 3D Gaussians. Additionally, an optional non-rigid deformation network can be used to further improve the performance.}
    \label{fig:overview}
\end{figure*}

This section initiates with a concise introduction to 3D Gaussian Splatting in Sec~\ref{sec:3D-GS}. 
Subsequently, in Sec.\ref{sec:SP}, we elaborate on how to apply a time-variant deformation network to the superpoints for predicting the rotation and translation to render images at any timestep. To fully exploit the \textit{As-Rigid-As-Possible} feature within one superpoint, our method also introduces a property reconstruction loss in Sec.\ref{sec:loss}. We also illustrate how to aggregate 3D Gaussians into superpoints using a learnable association matrix in this section. Moreover, some details of optimization and inference are explained in Sec.~\ref{sec:opt}. Finally, our method can support the plugin of an optional non-rigid deformation network, we clarify this in Sec.~\ref{sec:non-rigid}.
An overview of our method is illustrated in Fig. \ref{fig:overview}. 

\subsection{Preliminary: 3D Gaussian Splatting} \label{sec:3D-GS}
3D Gaussian Splatting(3D-GS)~\cite{3D-GS} propose a novel point-like scene representation, referred to as 3D Gaussians $\mathcal{G}=\{G_i: \gsP_i, \bm{s}_i, \bm{q}_i, \bm{\sigma}_i, \bm{h}_i\}$. 
Each 3D Gaussian $G_i$ is defined by a 3D covariance matrix $\mathbf{\Sigma}_i$ in world space~\cite{Zwicker2001EWAVS} and a center location $\gsP_i$, following the expression:
\begin{equation} \label{eq:gaussian}
    G_i(\bm{x}) = \exp\left(-\frac{1}{2}(\bm{x}-\gsP_i)^{\top}\mathbf{\Sigma}_i^{-1}(\bm{x}-\gsP_i)\right).
\end{equation}
For differentiable optimization, the covariance matrix $\mathbf{\Sigma}_i$ can be break down into a scaling matrix $\mathbf{S}_i$ and a rotation matrix $\gsR_i$, i.e., $\mathbf{\Sigma}_i = \gsR_i\mathbf{S}_i\mathbf{S}_i^{\top}\gsR_i^{\top}$, where $\mathbf{S}_i$ is represented by a 3D vector $\bm{s}_i$ and $\gsR_i$ is represented by a quaternion $\bm{q}_i \in \mathbf{SO}(3)$.

In the process of rendering a 2D image, 3D-GS projects 3D Gaussians onto a 2D image plane using the EWA Splatting algorithm~\cite{Zwicker2001SurfaceS}. The corresponding 2D Gaussian, defined by a covariance matrix $\mathbf{\Sigma}'$ in camera coordinates centered at $\gsP'$, is calculated as follows:
\begin{equation}
    \mathbf{\Sigma}' = \mathbf{J}\mathbf{W}\mathbf{\Sigma}\mathbf{W}^{\top}\mathbf{J}^{\top}, \quad \gsP' = \mathbf{J}\mathbf{W}\gsP,
\end{equation}
where $\mathbf{W}$ is the world-to-camera transformation matrix, and $\mathbf{J}$ is the Jacobian matrix of the affine approximation of the projective transformation. After sorting 3D Gaussians by depth, 3D-GS renders the image using volumetric rendering~\cite{VolumeRendering} (\ie $\alpha$-blending). The color $C(\bm{p})$ of pixel $\bm{p}$ is computed through blending $P$ ordered 2D Gaussians, as expressed by:
\begin{equation}
    \begin{split}
    C(\bm{p}) & = \sum_{i=1}^{P}\bm{c}_i\alpha_i \prod_{j=1}^{i-1}(1-\alpha_j), \\
    \alpha_i & = \bm{\sigma}_i \exp\left(-\frac{1}{2}(\bm{p}-\gsP'_i)^{\top}\mathbf{\Sigma}'_i(\bm{p}-\gsP'_i)\right),
    \end{split}
\end{equation}
where $\sigma_i$ represents the opacity of each 3D Gaussian, $\bm{c}_i$ is the RGB color computed using the spherical harmonics coefficients $\bm{h}_i$ of the 3D Gaussian and the view direction.

To optimize a static scene and facilitate real-time rendering, 3D-GS introduced a fast differentiable rasterizer and a training strategy that adaptively controls 3D Gaussians. Further details can be found in 3D-GS~\cite{3D-GS}, and the loss function utilized by 3D-GS is $\loss_1$ combined with a D-SSIM term:
\begin{equation}
    \loss_{img} = (1-\lambda)\loss_{1} + \lambda \loss_{\mathrm{D-SSIM}},
\end{equation}
where $\lambda$ is set to $0.2$.

\subsection{Superpoint Gaussian Splatting} \label{sec:SP}

It is evident that 3D-GS is suitable solely for representing static scenes. 
Therefore, when confronted with a monocular/multi-view video capturing a dynamic scene, we opt to learn 3D Gaussians in a canonical space and the deformation of each 3D Gaussian across the temporal domain under the guidance of aggregated superpoints. Since we assume there are only rigid transformations for every single 3D Gaussian, only the center location $\gsP_i$ and rotation matrix $\gsR_i$ of a 3D Gaussian will vary with time, while other attributes (e.g., opacity $\bm{\sigma}_i$, scaling vector $\bm{s}_i$, and spherical harmonics coefficients $\bm{h}_i$) remain invariant.

To model dynamic scene, we divide 3D Gaussians into  $M$ superpoints $\{\SP_j\}_{j=1}^{M}$ (\ie disjoint sets). Each superpoint $\SP_j$ contains several 3D Gaussians, while each 3D Gaussian has only one correspondent superpoint $\SP_j$. Following the principle of \textit{As-Rigid-As-Possible}, 3D Gaussians in the same superpoint $\SP_j$ should have similar deformation, which can represented by relative translation $\spT_j$ and rotation $\spR_j$ based on their center locations and rotation matrices in the canonical space.
Therefore, the center location $\gsP_i^t$ and rotation matrix $\gsR_i^t$ of the $i$-th 3D Gaussian at time $t$ will be:
\begin{equation} \label{transformation}
    \gsP_i^t = \spR^t_j \gsP^c_i  + \spT^t_j, \gsR_i^t = \spR^t_j \gsR_i^c.
\end{equation}

So as to predict the relative translation $\spT_j^t$ and rotation $\spR_j^t$ of the $j$-th superpoint at time $t$, we directly employ a deformation neural network $\mathcal{F}$ that takes the timestep $t$ and canonical position  $\spP^c_j$ of the $j$-th superpoint as input, and outputs the relative transformations of superpoints with respect to the canonical space:
\begin{equation} \label{eq:mlp-F}
    (\spR_j^t, \spT_j^t) = \mathcal{F}(\gamma(\spP^c_j), \gamma(t)),
\end{equation}
where $\gamma$ denotes the positional encoding:
\begin{equation}
    \gamma(x) = (\sin(2^k x), \cos(2^k x))_{k=0}^{L-1},
\end{equation}
In our experiments, we set $L=10$ for $\gamma(\spP^c_j)$ and $L=6$ for $\gamma(t)$.

During inference, to further decrease the rendering time, we can pre-compute the relative translation $\spT^t$ and rotation $\spR^t$ of superpoints predicted by the deformation network $\mathcal{F}$ at all timesteps. When rendering novel view images at a new timestep $t$ in the training set, the deformation of $j$-th superpoint can be calculated through interpolation:
\begin{equation} \label{eq:interp}
    \begin{split}    
    \spT^t_j & = (1-w)\spT^{t_1}_j + w\spT^{t_2}_j, \\
    \spR^t_j & = (1-w)\spR^{t_1}_j + w\spR^{t_2}_j, 
    \end{split}
\end{equation}
where the linear interpolation weight $w = (t-t_1)/(t_2-t_1)$, and $t_1$ and $t_2$ are the two nearest timesteps in the training dataset.

\subsection{Property Reconstruction Loss} \label{sec:loss}

The key insight of superpixels/superpoints lies in that pixels/points with similar properties should be aggregated into one group.
Following this idea, given an arbitrary timestep $t$, properties including the position $\spP^t$, the relative translation $\spT^t$ and relative rotation $\spR^t$ should be similar within one superpoint.
We utilize a learnable association matrix $\Asso\inR{P\times M}$ to establish the connection between 3D Gaussians and superpoints, where $P$ is the number of 3D Gaussians and $M$ is the number of superpoints.
Notably, only $K$ nearest superpoints of each Gaussian should be considered. 
Therefore, the associated probability $a_{ij}$ between Gaussian $G_i$ and superpoint $\SP_j$ can be calculated as:
\begin{equation}
    a_{ij} = \begin{cases}
      \displaystyle  \frac{\exp(\Asso_{ij})}{\sum_{j\in \knn_i}{\exp(\Asso_{ij})}}, & j \in \knn_i, \\
        0, & \text{otherwise},
    \end{cases}
\end{equation}
where $\knn_i$ is the set of $K$-nearest superpoints for the $i$-th Gaussian in the canonical space.

With the associated probability $a_{ij}$, the properties $\bm{u}_j \in \{\spP^t_j, \spR_j^t, \spT_j^t\}$ of $j$-th superpoint can be reconstructed from the properties of Gaussians:
\begin{equation} \label{eq:g2sp}
\begin{split}
   \gRsp: \bm{u}_j  & =\sum_{i \mid j \in \knn_i} \bar{a}_{ij} \bm{v}_i, \\
   \mbox{where~} \bar{a}_{ij} & = \frac{a_{ij}}{\displaystyle \sum_{i \mid  j \in \knn_i}{a_{ij}}},
\end{split}
\end{equation}
where $\bm{v}_i$ denotes the properties of $i$-th Gaussian, and ${i \mid j \in \knn_i}$ means all 3D Gaussians $i$ with the $j$-th superpoint in $\knn_i$.
It is noteworthy that the relative rotation $\spR_i^t$ is represented by Lie algebra $\se3$, which enables linear rotation interpolation. 
On the other hand, the properties $\bm{v}_i$ of the $i$-th Gaussian can also be reconstructed through adjacent superpoints:
\begin{equation} \label{eq:sp2g}
   \spRg: \bm{v}_i = \sum_{j \in \knn_i} a_{ij} \bm{u}_j.
\end{equation}

Ultimately, the property reconstruction loss is employed to ensure the consistency between the original properties $\bm{v}_i$ of Gaussians and the reconstructed properties $\bm{v}'_i$:
\begin{equation} \label{eq:pr-loss}
     \loss_{\bm{v}} = \frac{1}{P} \sum_{i}{\|\bm{v}_i, \bm{v}'_i\|},
 \end{equation}
 where $\bm{v}' = R_{sp\to g} (R_{g\to sp}(\bm{v}))$, and $\|\cdot, \cdot\|$ denotes the mean square error(MSE). And the more similar the Gaussian properties within the same superpoint are, the smaller this loss will be, thereby fully exploiting the \textit{As-Rigid-As-Possible} feature.

Furthermore, the corresponding superpoint $\SP_j$ of Gaussian $G_i$ is the superpoint with the highest association probability:
\begin{equation}
    j^* = \argmax_{j \in \knn_i}a_{ij}.
\end{equation}
It is noteworthy that $j^*$ is the same as the $j$ in Eq. \ref{transformation}.
 
\subsection{Optimization and Inference} \label{sec:opt}
The computation of the overall loss function is:
\begin{equation}
    \loss = \loss_{img} + \sum_{\bm{v} \in \{\gsP^t, \spR^t, \spT^t\}} \lambda_{\bm{v}} \loss_{\bm{v}},
\end{equation}
where $\lambda_{\bm{v}}$ represents hyper-parameters controlling the weights, with $\lambda_{\gsP^t} = 10^{-3}$ and $\lambda_{\spR^t} = \lambda_{\spT^t}=1$.

We implement our SP-GS with PyTorch, and $\mathcal{F}$ is a 8-layer MLPs with 256 hidden neurons. The network is trained for a total of 40k iterations, with the initial 3k iterations training without the deformation network $\mathcal{F}$ as a warm-up process to achieve relatively stable positions and shapes.
3D Gaussians in the canonical space will be initialized after the warm-up training, and for the initialization of superpoints, $M$ Gaussians are sampled using the farthest point sampling algorithm, and the canonical positions $\spP^c$ of superpoints are equal to the centers of the sampled Gaussians.
Moreover, the $A_{ij}$ of the learnable association matrix $\Asso$ will be initialized as 0.9 if the $j$-th superpoint is initialized with the $i$-th 3D Gaussian. Otherwise, $A_{ij}$ will be initialized as 0.1.
Before each iteration, we calculate the canonical position of superpoints with Eq. \ref{eq:g2sp}.

The Adam optimizer~\cite{Adam} is employed to optimize our models. For 3D Gaussians, the training strategies are the same as those of 3D-GS unless stated otherwise. For the learnable parameters of $\mathcal{F}$, the learning rate undergoes exponential decay, ranging from 1e-3 to 1e-5. The values for Adam's $\beta$ are set to (0.9, 0.999).

\subsection{Optional Non-Rigid Deformation Network}\label{sec:non-rigid}
Given the potential existence of non-rigid deformation in a dynamic scene, another optional non-rigid deformation network $\mathcal{G}$ is employed to learn the non-rigid deformation of each Gaussian for time $t$:
\begin{equation}
    (\hat{\spR}_i^t, \hat{\spT}_i^t) = \mathcal{G}(\gamma(\gsP_i^t)), \gamma(\bm{t})).
\end{equation}
By combining rigid motion with non-rigid deformation, the final center $\gsP^t_i$ and rotation matrix $\gsR^t_i$ of Gaussian $G_i$ can be computed as below:
\begin{equation}
    \begin{split}
        \gsP_i^t & = \hat{\spR}_i^t (\spR_j^t \gsP^c_i + \spT_j^t) + \hat{\spT}_i^t, \\
        \gsR_i^t & = \hat{\spR}_i^t \spR_j^t \gsR^c_i.
    \end{split}
\end{equation}
For the version incorporating the non-rigid deformation network $\mathcal{G}$ (abbreviated as SP-GS+NG), the model is initialized with the pretrained model from the version with only $\mathcal{F}$ and trained for 20k iterations using the loss $\loss_{img}$. Besides, $\mathcal{G}$ is a 3-layer MLPs with 64 hidden neurons.

%% file: sec/4_experiment.tex
\section{Experiment}
\newcommand{\citeInTab}[1]{}

We demonstrate the efficiency and effectiveness of our proposed approach with experiments on three datasets: the synthetic dataset D-NeRF~\cite{D-NeRF} with 8 scenes, the real-world dataset HyperNeRF ~\cite{Nerfies} and NeRF-DS~\cite{NeRF-DS}. 
For all experiments, we report the following metrics: PSNR, SSIM~\cite{SSIM}, MS-SSIM,  LPIPS~\cite{LPIPS}, size (rendering resolution), and FPS (rendering speed). 
All experiments are conducted on one NVIDIA V100 GPU with 32GB memory.

Regarding the baselines, we compare our method against the state-of-the-art methods that are the most relevant to our work, including: D-NeRF~\cite{D-NeRF}, TiNeuVox~\cite{TiNeuVox}, Tensor4D~\cite{Tensor4D}, K-Palne~\cite{KPlanes},  HexPlane~\cite{HexPlane}, TI-DNeRF~\cite{TI-DNeRF}, NeRFPlayer~\cite{NeRFPlayer}, 4D-GS~\cite{RealTime4DGS}, Deformable 3D GS(D-3D-GS)~\cite{Deformable3DGS} and original 3D Gaussians (3D-GS).

\subsection{Synthetic Dataset}

\begin{table}[t]
    \centering
    \caption{Quantitative comparison on D-NeRF~\cite{D-NeRF}. The \colorTop[1]{best} and \colorTop[2]{second best} results are highlighted. `-' denotes that the metric is not reported in their works. Lego is excluded.}
    \label{tab:D-NeRF-avg}
    \resizebox{1.0\linewidth}{!}{
    \setlength{\tabcolsep}{2pt}
    \begin{tabular}{c|*{5}{c}}
        \toprule
        Methods & PSNR$\uparrow$ & SSIM$\uparrow$ & LPIPS$\downarrow$ & Size & FPS$\uparrow$ \\ \midrule
        D-NeRF\citeInTab{DNeRF} & 31.14 &  0.9761 & 0.0464 & $400\times 400$ & $<1$   \\
        TiNeuVox-B\citeInTab{TiNeuVox} & 32.74  & 0.9715 & 0.0495 & $400\times 400$ & $\sim$ 1.5 \\
        Tensor4D\citeInTab{Tensor4D} & 27.44 &  0.9421  &  0.0569 & $400\times 400$ & -\\
        KPlanes\citeInTab{KPlanes} & 31.41 & 0.9699  & 0.0470 & $400\times 400$ & $\sim$ 0.12 \\
        HexPlane-Slim\citeInTab{HexPlane} & 32.97 & 0.9750 & 0.0346 & $400\times 400$ & 4 \\
        Ti-DNeRF\citeInTab{TI-DNeRF} & 32.69 & 0.9746 & 0.358 & $400\times 400$ & -  \\
        \midrule
        3D-GS\citeInTab{3D-GS} & 23.39 & 0.9293 & 0.0867 & $800\times 800$ & \colorTop[2]{184.21} \\
        4D-GS\citeInTab{4D-GS} & 35.31  & 0.9841  & \colorTop[2]{0.0148}  & $800\times 800$ & 143.69 \\
        D-3D-GS\citeInTab{Deformable3DGS} & \colorTop[1]{40.23} & \colorTop[1]{0.9914} & \colorTop[1]{0.0066} &
        $800\times 800$ & 45.05 \\
        \textbf{SP-GS(ours)} & 37.98  & 0.9876  & 0.0185 & $800\times 800$ & \colorTop[1]{227.25} \\
        \textbf{SP-GS+NG(ours)} & \colorTop[2]{38.28}  & \colorTop[2]{0.9877}   & 0.0152 & $800\times 800$ & 119.35  \\
        \bottomrule
    \end{tabular}
    }
\end{table}

\begin{figure*}[t]
    \centering
    \includegraphics[width=0.9\linewidth]{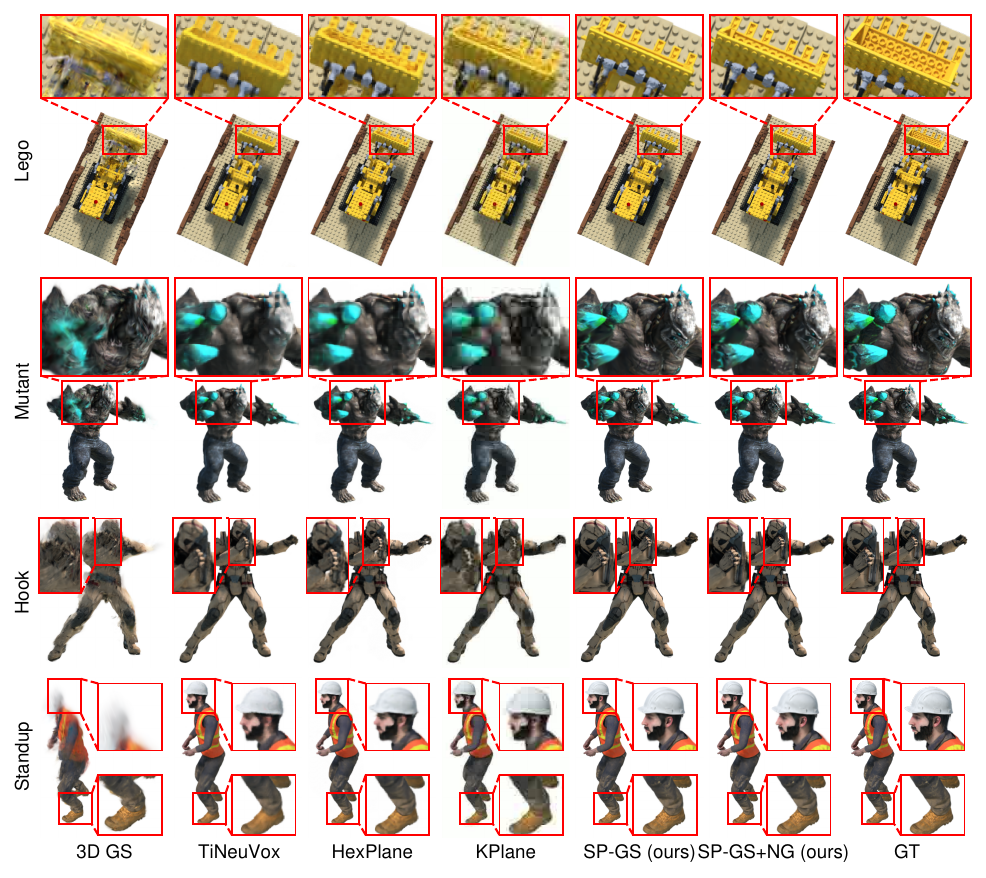}
    \caption{Qualitative comparisons of baselines and our method on D-NeRF~\cite{D-NeRF}.}
    \label{fig:cmp-D-NeRF}
\end{figure*}

The D-NeRF dataset consists of 8 videos, each containing 50-200 frames. The frames together with camera pose serve as the training data, while test views are taken from novel views.
Quantitative and qualitative results are shown in Tab.\ref{tab:D-NeRF-avg} and Fig.\ref{fig:cmp-D-NeRF}. Though rendered at a resolution of $800\times 800$, we achieved a much higher FPS than previous non-Gaussian-Splatting based methods. As for D-3D-GS, they directly apply a deformation network to every single 3D Gaussian for a higher visual quality, leading to a much lower FPS than ours. We achieve superior or comparable results against previous state-of-the-art methods in terms of all metrics. It's noteworthy that Lego is excluded while calculating the metrics as we observed a discrepancy in all methods. Please refer to Fig. \ref{fig:cmp-D-NeRF} for a visualized results.
Per-scene comparisons are provided in Appendix \ref{sec-ap:DNeRF-results}.

\subsection{Real-World Dataset}

\begin{figure*}[t]
    \centering
    \setlength{\tabcolsep}{1pt}
    \begin{tabular}{*{7}{c}}
    \includegraphics[width=0.14\linewidth]{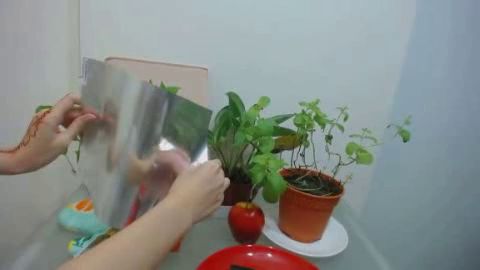} & 
    \includegraphics[width=0.14\linewidth]{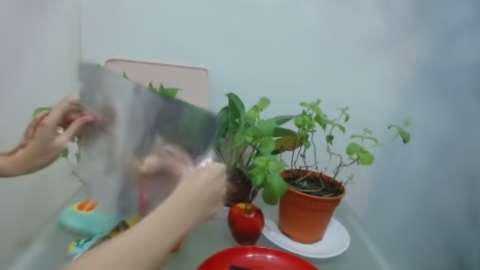} &
    \includegraphics[width=0.14\linewidth]{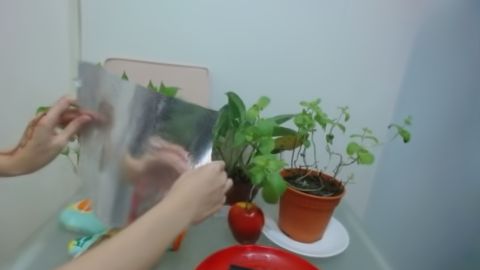} &
    \includegraphics[width=0.14\linewidth]{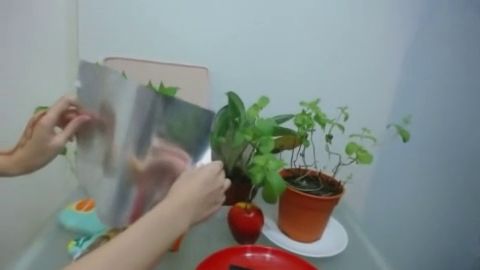} &
    \includegraphics[width=0.14\linewidth]{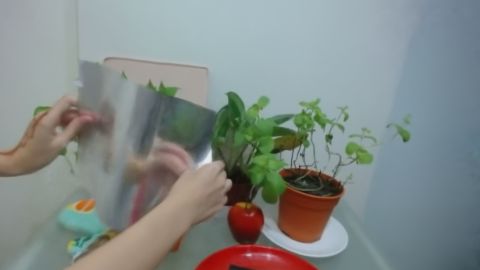} &
    \includegraphics[width=0.14\linewidth]{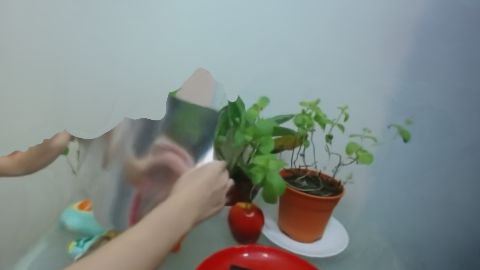} &
    \includegraphics[width=0.14\linewidth]{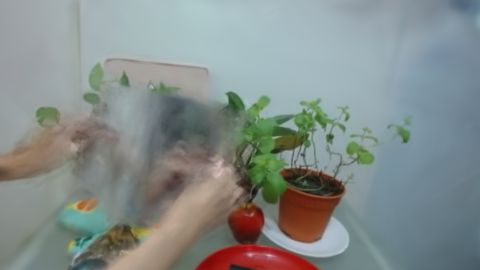} 
    \\
    GT & \textbf{SP-GS(ours)} & D-3D-GS & NeRF-DS & HyperNeRF & TiNeuVov & 3D-GS\\ 
    \end{tabular}
    \caption{Qualitative comparisons of baselines and our method on NeRF-DS dataset~\cite{NeRF-DS}.}
    \label{fig:cmp-NeRF-DS}
\end{figure*}

\begin{figure*}[t]
    \centering
    \setlength{\tabcolsep}{1pt}
    \begin{tabular}{*{8}{c}}
    \includegraphics[width=0.12\linewidth]{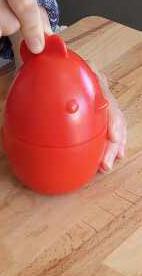} & 
    \includegraphics[width=0.12\linewidth]{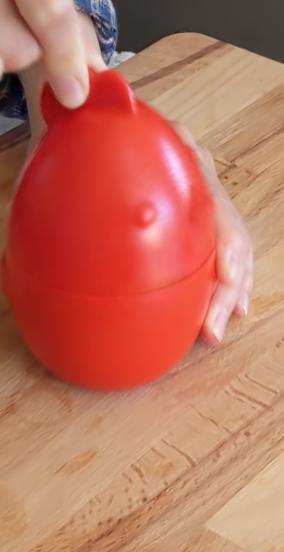} & 
    \includegraphics[width=0.12\linewidth]{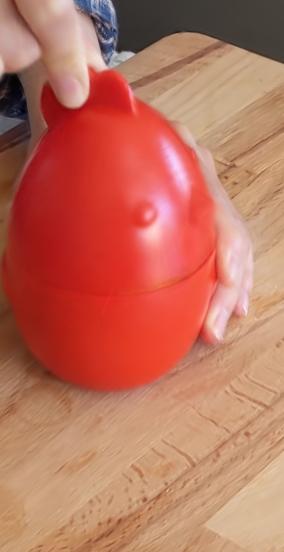} & 
    \includegraphics[width=0.12\linewidth]{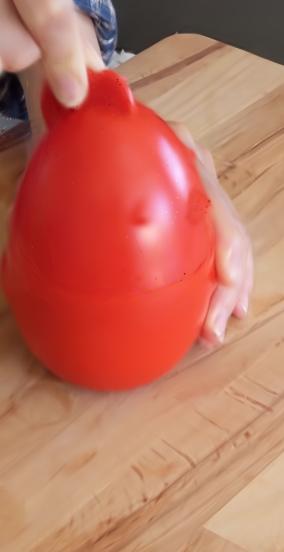} & 
    \includegraphics[width=0.12\linewidth]{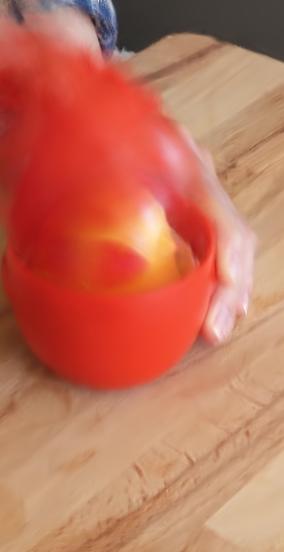} & 
    \includegraphics[width=0.12\linewidth]{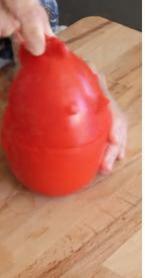} & 
    \includegraphics[width=0.12\linewidth]{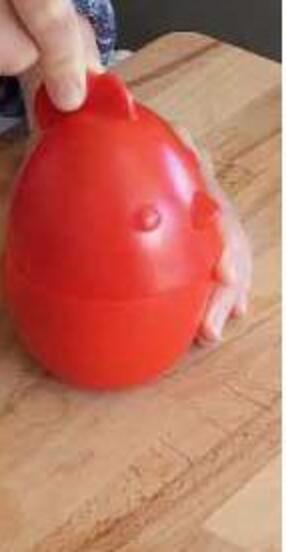} & 
    \includegraphics[width=0.12\linewidth]{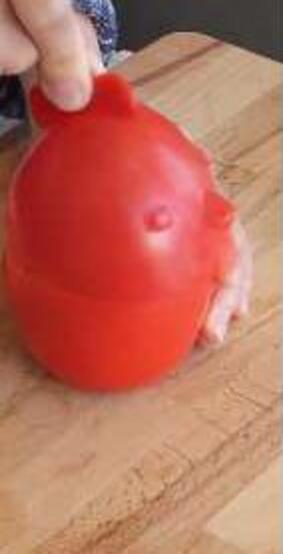} 
    \\
    GT & \textbf{SP-GS (ours)} & \textbf{SP-GS+NG} & 4D-GS & 3D-GS &  NeRFPlayer &  HyperNeRF & Nerfies 
    \end{tabular}
    \caption{Qualitative comparisons of baselines and our method on HyperNeRF dataset\cite{HyperNeRF}.}
    \label{fig:cmp-HyperNeRF}
\end{figure*}

The HyperNeRF dataset~\cite{Nerfies} and NeRF-DS~\cite{NeRF-DS} serve as two real-world benchmark dataset captured using either one or two cameras. For a fair comparison with previous methods, we use the same \textit{vrig} scenes, a subset of the HyperNeRF dataset.
Quantitative and qualitative results of the HyperNeRF dataset are shown in Tab.\ref{tab:HyperNeRF-avg} and Fig.\ref{fig:cmp-HyperNeRF}, while results of NeRF-DS are shown in Tab. \ref{tab:NeRF-DS} and Fig. \ref{fig:cmp-NeRF-DS}. As shown in Tab. \ref{tab:HyperNeRF-avg} and Tab. \ref{tab:NeRF-DS}, our method outperforms baselines by a large margin in terms of FPS while achieving a superior or comparable visual quality. As shown in Fig.\ref{fig:cmp-HyperNeRF}, our results exhibit notably higher visual quality, particularly in the hand area.

\begin{table}[t]
    \centering
    \caption{Quantitative comparison on HyperNeRF dataset~\cite{HyperNeRF}. The \colorTop[1]{best} and \colorTop[2]{second best} results are highlighted. `-' denotes that the metric is not reported in their works.}
    \label{tab:HyperNeRF-avg}
    \resizebox{\linewidth}{!}{
    \setlength{\tabcolsep}{2pt}
    \begin{tabular}{*{6}{c}}
        \toprule
        Methods  & PSNR$\uparrow$ & MS-SSIM  $\uparrow$ & LPIPS  $\downarrow$ & Size &  FPS $\uparrow$  \\ \midrule
        Nerfies\citeInTab{Nerfies} & 22.2 & 0.803 & - & $536\times 960$ & $< 1$ \\
        HyperNeRF\citeInTab{HyperNeRF} & 22.4 & 0.814 & -&$536\times 960$ & $< 1$  \\
        TiNeuVox-S \citeInTab{TiNeuVox} & 23.4 & 0.813 & -& $536\times 960$ & $< 1$ \\
        TiNeuVox-B \citeInTab{TiNeuVox}  & 24.3 & 0.837 & - & $536\times 960$ &  $< 1$  \\
        TI-DNeRF\citeInTab{TI-DNeRF} & 24.35 & \colorTop[2]{0.866} && $536\times 960$ & $< 1$\\
        NeRFPlayer\citeInTab{NeRFPlayer} & 23.7 & 0.803 & - & $536\times 960$ & $< 1$\\
        \midrule
        3D-GS \citeInTab{3D-GS} & 20.26 &  0.6569 & 0.3418 &$536\times 960$ & \colorTop[2]{71}  \\
        4D-GS \citeInTab{4D-GS}  &  25.02 &  0.8377 &  0.2915
& $536\times 960$ & 66.21   \\
        \textbf{SP-GS (ours)} & \colorTop[2]{25.61} & 0.8404  & \colorTop[2]{0.2073} &$536\times 960$ & \colorTop[1]{117.86} \\
        \textbf{SP-GS+NG (ours)} & \colorTop[1]{26.78} & \colorTop[1]{0.8920} & \colorTop[1]{0.1805} &$536\times 960$ & 51.51\\
        \bottomrule
    \end{tabular}
    }
\end{table}

\begin{table}[t]
    \centering
    \caption{Quantitative comparison on NeRF-DS~\cite{NeRF-DS}. The rendering size is $480 \times 270$.
    }
    \label{tab:NeRF-DS}
    \resizebox{\linewidth}{!}{
    \setlength{\tabcolsep}{2pt}
    \begin{tabular}{c*{5}{c}}
    \toprule
      Methods   & PSNR$\uparrow$ & SSIM$\uparrow$ & LPIPS(VGG)$\downarrow$ & FPS$\uparrow$ \\ \midrule
    TiNeuVox       & 21.61 & 0.8241 & 0.3195 & - \\
    HyperNeRF      & 23.45 & 0.8488 & 0.2002 & - \\
    NeRF-DS        & 23.60 & 0.8494 & 0.1816 & - \\ 
    \midrule
    3D-GS          & 20.29 & 0.7816 & 0.2920 & 185.43 \\ 
    D-3D-GS & 24.11 & 0.8525 & 0.1769 & 15.27 \\
\textbf{SP-GS(ours)} & 23.15 & 0.8335 & 0.2062 & 251.70 \\
\textbf{SP-GS+NG(ours)} & 23.33 & 0.8362  & 0.2084 & 66.13\\
    \bottomrule
    \end{tabular}
    }
\end{table}

\subsection{Ablation Study}
In our paper, we introduce two hyperparameters: the number of superpoints and the number of nearest neighborhoods $\mathcal{N}_i$. We conduct experiments on testing how sensitive our method is to the variation of the two hyperparameters. 
Tab. \ref{tab:as_num} shows the performance of our approach when varying these hyperparameters, and our method appears to be robust under all these variations. 
Besides, we introduce property reconstruction loss to facilitate grouping similar Gaussians together. Tab. \ref{tab:property-loss} demonstrates that property reconstruction loss can improve rendering quality.
We provide more ablations on property reconstruction loss in Appendix.\ref{sec-ap:more-exp}.

\subsection{Visualization of 3D Gaussians and Superpoints}
\begin{figure}[t]
    \centering
    \subfigure[]{\includegraphics[width=0.24\linewidth, trim=50 50 50 50,clip]{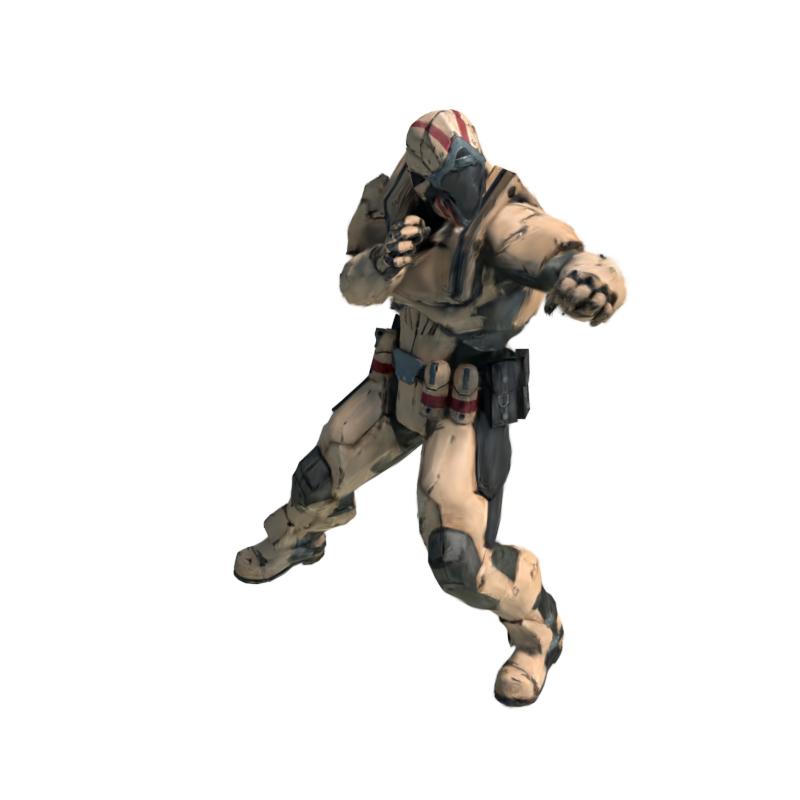}}
    \subfigure[]{\includegraphics[width=0.24\linewidth, trim=50 50 50 50,clip]{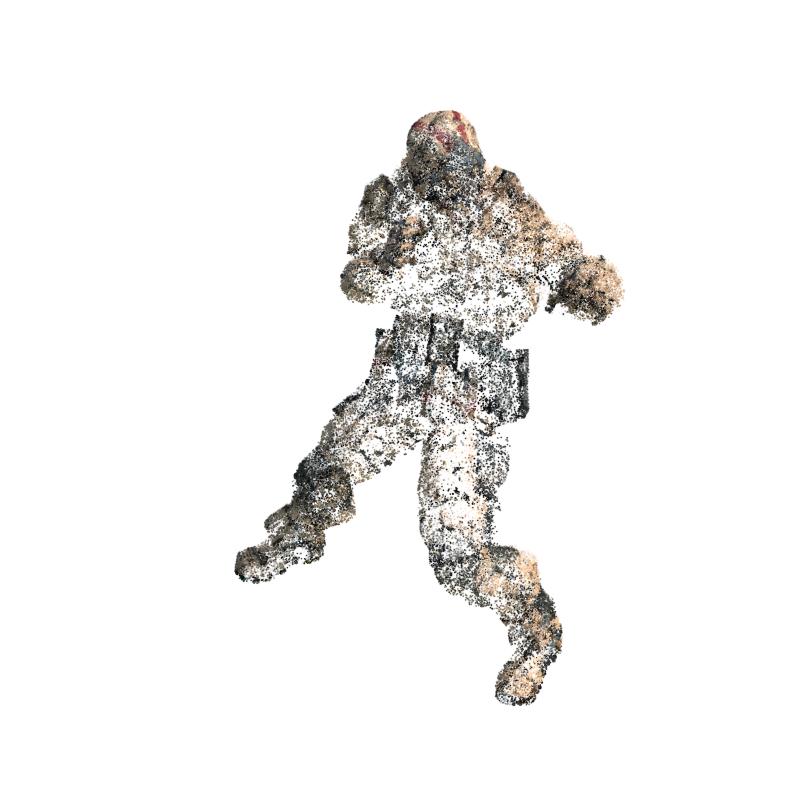}}
    \subfigure[]{\includegraphics[width=0.24\linewidth, trim=50 50 50 50,clip]{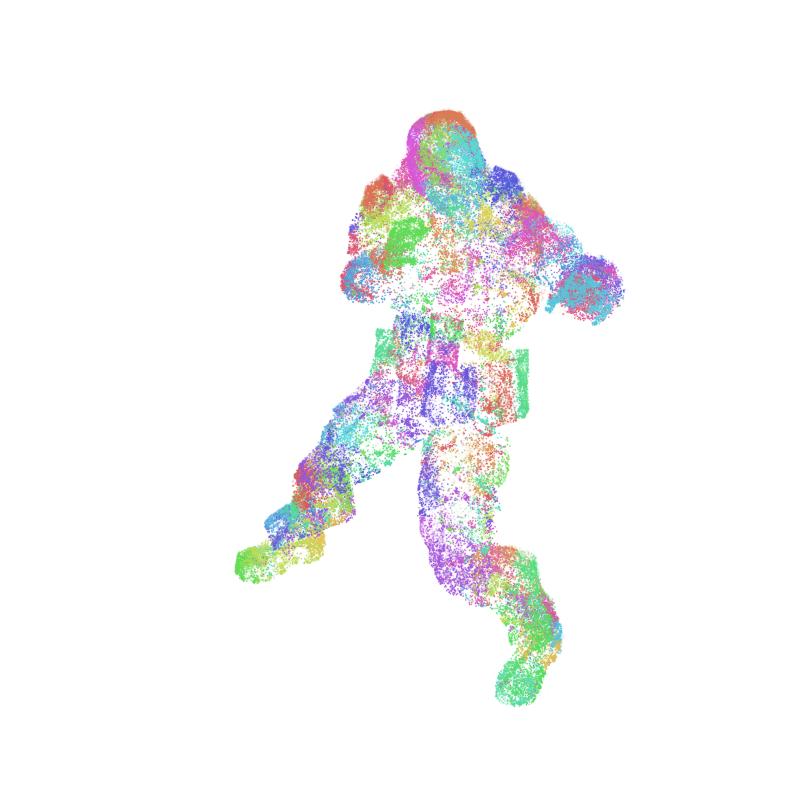}}
    \subfigure[]{\includegraphics[width=0.24\linewidth, trim=50 50 50 50,clip]{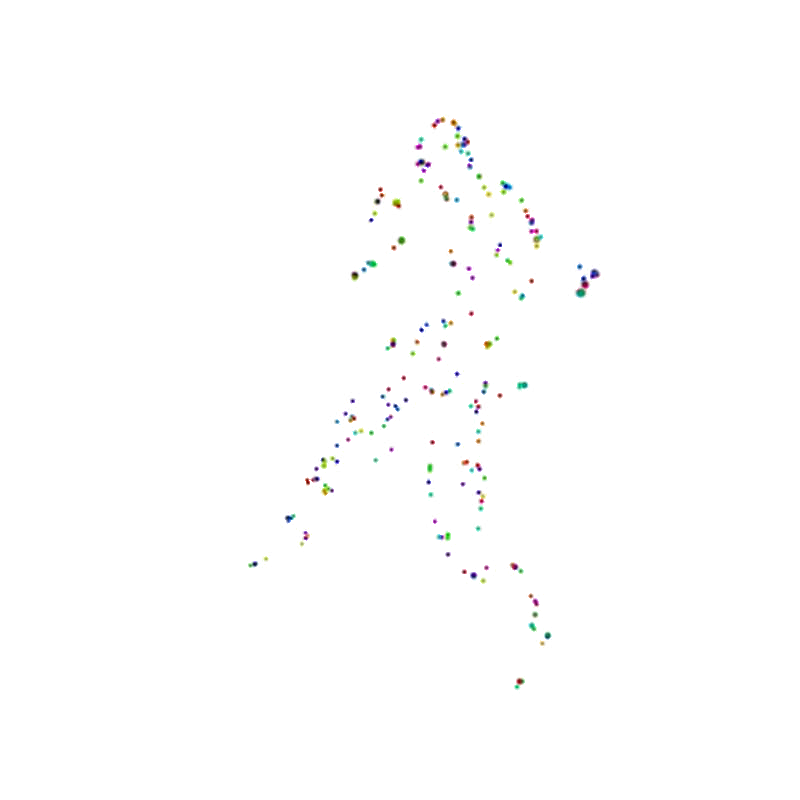}}
    \caption{We visualize 3D Gaussians and superpoints by simply coloring the correspondent points. (a) the rendered image. (b) 3D Gaussians with its original color. (c) 3D Gaussians colored by superpoints, which means 3D Gaussians in one superpoint will be the same color. (d) the superpoints.}
    \label{fig:vis}
\end{figure}

As depicted in Fig. \ref{fig:vis}, we provide a visualized result of the 3D Gaussians and superpoints for the hook scene of D-NeRF~\cite{D-NeRF}. Notably, we observe that the superpoints are uniformly distributed in the space while nearby 3D Gaussians will be aggregated into one superpoint. 
For a more intuitive understanding, readers can refer to our \href{https://dnvtmf.github.io/SP_GS.github.io/}{project page} for more videos.

\begin{table}[t]
    \centering
    \caption{Ablation Study for the number of superpoints (\#sp) and nearest neighborhoods (\#knn) on D-NeRF dataset.}
    \label{tab:as_num}
    \resizebox{0.9\linewidth}{!}{
        \begin{tabular}{c*{6}{c}}
            \toprule
        \#sp & 50 & 100 & 200 & 300 & 400 & 500  \\
        \midrule
        PSNR$\uparrow$ & 35.69 & 36.00 & 36.31 & 36.43 & 36.36 & 36.52\\ \bottomrule
        \#knn & 1 & 2 & 3 &4 &5&6\\ \midrule
        
        PSNR$\uparrow$ &36.24 & 36.11 & 36.09 & 36.09 & 36.30 & 36.23 \\ \bottomrule
        \end{tabular}
    }
\end{table}

\begin{table}[t]
    \centering
    \caption{Ablation study for property reconstruction loss on D-NeRF dataset.}
    \label{tab:property-loss}
    \begin{tabular}{cccc}
        \toprule
        Method & PSNR$\uparrow$ & SSIM$\uparrow$ & LPIPS$\downarrow$ \\ \midrule
        w/o loss & 37.59 & 0.9868 &  0.0172\\
w loss	& 37.98 & 0.9876 & 0.0164  \\ \bottomrule     
    \end{tabular}
\end{table}

\begin{table}[t]
    \centering
    \caption{The results of distillation on ``As" scene of NeRF-DS dataset. We use D-3D-GS as teacher model, and use SP-GS as student model.}
    \label{tab:distill}
    \setlength{\tabcolsep}{1pt}
    \resizebox{\linewidth}{!}{
        \begin{tabular}{*{6}{c}}
            \toprule
            Method & PSNR$\uparrow$& MS-SSIM $\uparrow$ & LPIPS$\downarrow$ & FPS$\uparrow$  \\ \midrule
            teacher (D-3D-GS) & 26.15 & 0.8816 & 0.1829 & 20.65\\
            student (SP-GS) & 25.68 & 0.8811 & 0.1982 & 164.04 \\
            ours (SP-GS) & 	24.44 & 0.8626 & 0.2255 & 250.32 \\
            \bottomrule
        \end{tabular}
    }
\end{table}

\begin{figure}[thb]
    \centering
    \subfigure[Pose Estimation]{\label{fig:repose}\includegraphics[width=0.48\linewidth]{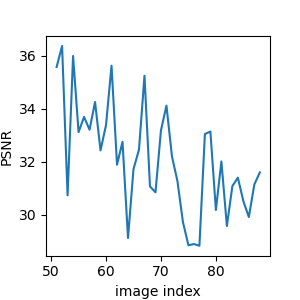}}
    \subfigure[Scene Editing]{\label{fig:edit} \includegraphics[width=0.48\linewidth]{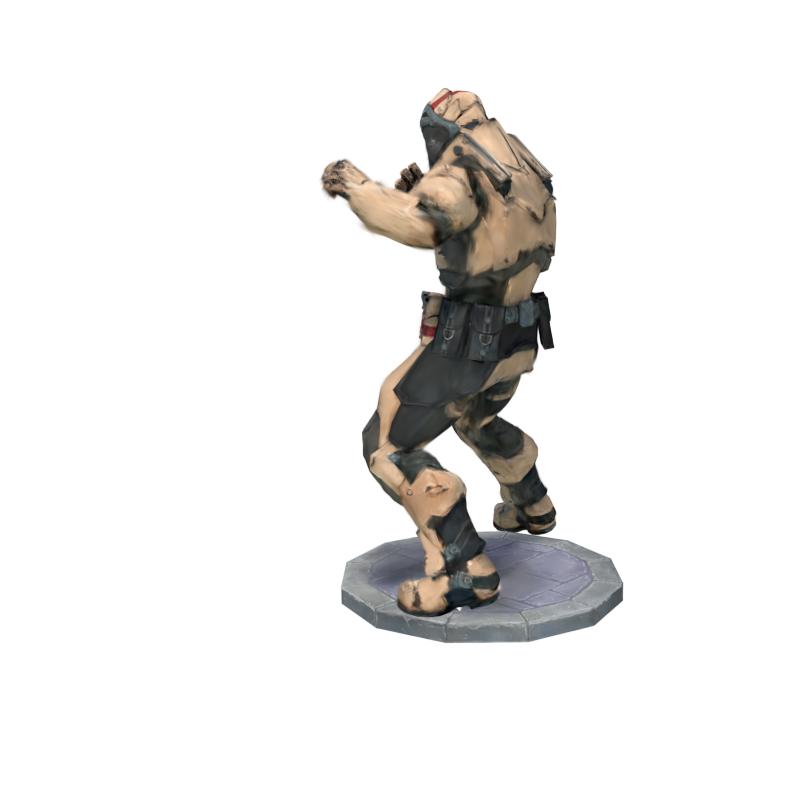}}
    \caption{(a) The PSNR results for pose estimation results on the jumpingjacks scene. (b) We add the base of the trex scene to the bottom of the hook scene.}
    \label{fig:app}
\end{figure}

%% file: sec/5_applications.tex
\section{Applications}
Thanks to the powerful representation of superpoints for dynamic scenes, our method is highly expandable and can facilitate various downstream applications. 
\subsection{Model Distillation}
In scenarios where a 3D-GS based model $\mathcal{H}$ exhibits superior performance, which also predicts the deformation over time, we can distill such a model into SP-GS to improve the visual quality.
The concept is straightforward: we can directly replicate the state of $\mathcal{H}$ at any given time as the canonical state of SP-GS. 
Subsequently, we optimize the association matrix $\Asso$, the superpoint deformation network $\mathcal{F}$, and optionally the non-rigid deformation network $\mathcal{G}$ by incorporating $\loss$ loss and mean square error(MSE) $ \loss_{err} =\sum_{\bm{u} \in \{\gsP, \gsR \}} \lambda_{\bm{u}} \sum_{i} \|\bm{u}^t_i, \bm{u}'^t_i \|$,
% \begin{equation}
%     \loss_{err} =\sum_{\bm{u} \in \{\gsP, \gsR \}} \lambda_{\bm{u}} \sum_{i} \|\bm{u}^t_i, \bm{u}'^t_i \|,
% \end{equation}
where $\bm{u}^t_i$  and $\bm{u}'^t_i$ are the properties of the teacher and the student respectively.
Tab. \ref{tab:distill} shows the quantitative results of distilling D-3D-GS model into SP-GS model on the ``As" scene of NeRF-DS dataset.
While D-3D-GS cannot achieve real-time rendering on V100 (20.65 FPS), our distillated student model can achieve significantly higher rendering speed (164.04 FPS). Therefore, model distillation provides a trade-off between visual quality and rendering speed, leaving users with more choices to meet their requirements.

\subsection{Pose Estimation}

Our SP-GS support estimating the 6-DoF pose of each superpoint for new images in the same scene.
%  without relying on the deformation network. 
To be specific, we can solely learn the translation and rotation for each superpoint with other parameters of SP-GS fixed. This can be potentially used in motion capture where novel view images are given and one wants to know the motion of each components (superpoints).
Experiments are conducted on the jumpingjacks scene of D-NeRF dataset~\cite{D-NeRF}. We first train the complete model (SP-GS) using the beginning 50 images. Subsequently, we initialize the learnable translation and rotation parameters of superpoints with the 50-th frame and directly optimize them with rendering loss using the Adam optimizer for 1000 iterations.
The program terminates upon completing the pose estimation for all images.
Fig. \ref{fig:repose} illustrates the change in PSNR across images 51-88 for the jumpingjacks scene.
Notably, it demonstrates a gradual decrease in PSNR.

\subsection{Scene Editing}
As depicted in Figure \ref{fig:edit}, scene editing tasks such as relocating parts between scenes or removing parts from a scene can be accomplished with ease. 
This capability is facilitated by the explicit 3D Gaussian representation, enabling relocation or deletion from the scene.
Our method further streamline the process since it is no longer necessary to manipulate over the 100,000 3D Gaussians. Moreover, our superpoints provide some extent of semantic meanings, enabling a reasonable editing of the scene.

%% file: sec/6_conclusions.tex
\section{Limitations}
Similar to 3D-GS, the reconstruction of real-world scenes requires sparse point clouds to initialize 3D scenes. However, it it challenging for software like COLMAP~\cite{colmap}, which is designed for static scenes, to initialize point clouds, resulting in diminished camera poses. Consequently, these issues may impede the convergence of our SP-GS to the expected results. We aim to address it in future work.

\section{Conclusions}
This paper introduces Superpoint Gaussian Splatting as a novel method for achieving real-time, high-quality rendering for dynamic scenes. Building upon 3D-GS, our approach involves grouping Gaussians with similar motions into superpoints, adding extremely small burden for Gaussians rasterization. 
Experimental results demonstrate the superior visual quality and rendering speed of our method while our framework can also support various downstream applications.

%% file: sec/7_other.tex
\section*{Acknowledgements}

This work is supported by the Sichuan Science and Technology Program (2023YFSY0008), China Tower-Peking University Joint Laboratory of Intelligent Society and Space Governance, National Natural Science Foundation of China (61632003, 61375022, 61403005), Grant SCITLAB-20017 of Intelligent Terminal Key Laboratory of SiChuan Province, Beijing Advanced Innovation Center for Intelligent Robots and Systems (2018IRS11), and PEKSenseTime Joint Laboratory of Machine Vision.

\section*{Impact Statement}
This paper presents work whose goal is to achieve photorealistic and real-time novel view synthesis for dynamic scenes.  
Therefore, we acknowledge that our approach can potentially be used to generate fake images or videos.
We firmly oppose the use of our research for disseminating false information or damaging reputations.

%% file: sec/X_suppl.tex
%mainfile: main.tex
\clearpage
\setcounter{section}{0}
\renewcommand\thesection{\Alph{section}}

The Appendix provides additional details concerning network training. Additional experimental results are also included, which are omitted from the main paper due to the limited space.

In Section \ref{sec-ap:details}, we elaborate on additional implementation details of our approach. Section \ref{sec-ap:results} presents the per-scene quantitative comparisons between our methods and baselines in D-NeRF~\cite{D-NeRF} dataset, HyperNeRF dataset~\cite{HyperNeRF}, NeRF-DS dataset~\cite{NeRF-DS} and the NVIDIA Dynamic Scene Dataset~\cite{Yoon2020NovelVS}. In Section \ref{sec-ap:more-exp}, we offer additional ablation studies of our method. 

\section{More Implementation Details} \label{sec-ap:details}

\begin{figure}[htb]
    \centering
    \includegraphics[width=\linewidth]{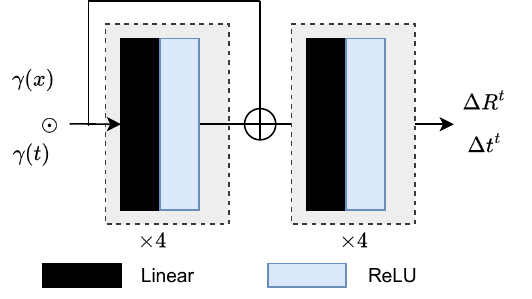}
    \caption{The architeture of deformation network $\mathcal{F}$. $\odot$ denotes concatenation operation.}
    \label{fig:MLP}
\end{figure} 
As depicted in Fig. \ref{fig:MLP}, superpoint deformation network $\mathcal{F}$ consists of 8-layer MLPs with a hidden dimension of 256, while the non-rigid deformation network $\mathcal{G}$ consists of 4-layer MLPs with a hidden dimension of 64 and ReLU activations. Upon initializing $\mathcal{F}$ and $\mathcal{G}$, the \textit{weights} and \textit{bias} of last layer are set to 0.

For the D-NeRF dataset, we initialize 3D Gaussians from random point clouds with 10k points. We apply densification and pruning to the 3D Gaussians every 100 iterations, starting from 600 iterations and stopping at 15k iterations. The opacity of 3D Gaussians is reset every 3k iterations until 15k iterations.

Concerning the real-world HyperNeRF Dataset, we utilize colmap to derive camera parameters and colored sparse point clouds. Densification and pruning of 3D Gaussians occur every 1000 iterations, starting from 1000 iterations and stopping at 15k iterations. The opacity of 3D Gaussians is reset every 6k iterations until 15k iterations.

\section{More about Real Time Rendering}
We compare the rendering speed between our SP-GS and other methods on different GPUs (\ie V100, TITAN Xp, GTX 1060). 
Results are shown in \cref{tab:speed-DNeRF}, \cref{tab:speed-HyperNeRF} and \cref{tab:speed-NeRF-DS}.
As can be seen from those tables, our SP-GS can achieve real-time rendering speed of complex scenes on devices with limited computing power (such as GTX 1060), while D-3D-GS cannot achieve real-time rendering.

\begin{table}[t]
    \centering
    \caption{Comparsion about rendering speed on different devices in D-NeRF dataset.}
    \label{tab:speed-DNeRF}
    \begin{tabular}{cccc}
        \toprule
        Method & V100 & TITAN Xp & GTX 1060 \\ \midrule
        NeRF-based & $< 1$ & $<1$ & $<1$ \\ 
        D-3D-GS &45.05 &	30.84 & 	13.44 \\
        4D-GS	& 143.69 &	134.16 &	95.01 \\
        \textbf{SP-GS (ours)} & \colorTop{227.25} & \colorTop{197.90} &	\colorTop{140.341}\\
        \bottomrule
    \end{tabular}
\end{table}

\begin{table}[t]
    \centering
    \caption{Comparsion about rendering speed on different devices in HyperNeRF dataset.}
    \label{tab:speed-HyperNeRF}
    \begin{tabular}{cccc}
        \toprule
        Method & V100 & TITAN Xp & GTX 1060 \\ \midrule
        NeRF-based & $< 1$ & $<1$ & $<1$ \\ 
        D-3D-GS & 4.87	& 4.71	& 2.00 \\
        4D-GS & 66.21& 	58.95& 	29.03\\
        \textbf{SP-GS (ours)} & \colorTop{117.86} & \colorTop{101.58} & \colorTop{56.06} \\
        \bottomrule
    \end{tabular}
\end{table}

\begin{table}[t]
    \centering
    \caption{Comparsion about rendering speed on different devices in NeRF-DS dataset.}
    \label{tab:speed-NeRF-DS}
    \begin{tabular}{cccc}
        \toprule
        Method & V100 & TITAN Xp & GTX 1060 \\ \midrule
        NeRF-based & $< 1$ & $<1$ & $<1$ \\ 
        D-3D-GS & 15.27	& 13.13 &	7.33 \\
        \textbf{SP-GS (ours)} & \colorTop{251.70} &	\colorTop{160.06} &	\colorTop{90.40}  \\
        \bottomrule
    \end{tabular}
\end{table}

\section{Per Scene Results} \label{sec-ap:results}

\subsection{D-NeRF Dataset} \label{sec-ap:DNeRF-results}
On the synthetic D-NeRF dataset, Tab.\ref{tab:DNeRF-pre-scene} illustrates per-scene quantitative comparisons in terms of PSNR, SSIM, and LPIPS (alex). Our approaches, SP-GS and SP-GS+NG, exhibit superior performance compared to non-Gaussian-Splatting based methods. In comparison to the concurrent work D-3D-GS~\cite{Deformable3DGS}, which employs heavy MLPs (8 layers, 256-D) on every single 3D Gaussian and cannot achieve real-time rendering, our results are slightly less favorable.

Tab. \ref{tab:D-NeRF-time} elucidates the number of the final 3D Gaussians (\textit{\#Gaussians}), the training time (\textit{train}), and the rendering speed (\textit{FPS}) for SP-GS and SP-GS+NG on the D-NeRF dataset. It is evident that a reduced number of Gaussians result in faster training time and rendering speed. Therefore, adjusting hyper-parameters of the "Adaptive Control of Gaussians" (e.g., densify and prune interval, threshold) is a possible way to achieve faster training time and rendering speed. 
It is worth noting that the training time of SP-GS+NG in table only includes the time of training the NG part, and does not include the training time for SP-GS.

\subsection{HyperNeRF Dataset}  \label{sec-ap:HyperNeRF-results}
For the HyperNeRF Dataset, Tab.\ref{tab:Hyper-pre-scene} reports the per scene results on \textit{vrig-broom}, \textit{vrig-3dprinter}, \textit{vrig-chichen} and \textit{vrig-peel-banana}. Tab. \ref{tab:Hyper-NeRF-time} elucidates the number of the final 3D Gaussians (\textit{\#Gaussians}), the training time (\textit{train}), and the rendering speed (\textit{FPS}) for SP-GS and SP-GS+NG on the HyperNeRF Dataset.

\begin{table}[thb]
    \centering
    \caption{Training time and rendering speed on the D-NeRF dataset.}
    \label{tab:D-NeRF-time}
    \resizebox{\linewidth}{!}{
    \begin{tabular}{cccccc}
        \toprule
        \multirow{2}{*}{scene}  & \multirow{2}{*}{\#Gaussians}   & \multicolumn{2}{c}{SP-GS} & \multicolumn{2}{c}{SP-GS+NG}  \\
        & & Train$\downarrow$ & FPS$\uparrow$ &Train$\downarrow$ & FPS$\uparrow$  \\ \midrule
        hellwarrior   & 38k  & 26m  & 249.93 & +8m  & 65.54  \\
        mutant        & 181k & 79m  & 210.42 & +9m  & 199.43 \\
        hook          & 135k & 60m  & 230.35 & +8m  & 107.82 \\
        bouncingballs & 41k  & 28m  & 181.77 & +9m  & 154.2  \\
        lego          & 264k & 113m & 168.89 & +16m & 45.32  \\
        trex          & 165k & 72m  & 186.65 & +11m & 91.76  \\ 
        standup       & 74k  & 38m  & 260.34 & +8m  & 155.6  \\
        jumpingjack   & 68k  & 39m  & 271.27 & +9m  & 61.13  \\
        \midrule
        average       & 120k & 57m  & 219.95 & +8.6m& 110.10 \\ \bottomrule
    \end{tabular}
    }
\end{table}

\begin{table}[thb]
    \centering
    \caption{The train time and rendering speed on the \textit{vrig} HyperNeRF Dataset.}
    \label{tab:Hyper-NeRF-time}
    \resizebox{\linewidth}{!}{
    \begin{tabular}{cccccc}
        \toprule
        \multirow{2}{*}{scene}  & \multirow{2}{*}{\#Gaussians}   & \multicolumn{2}{c}{SP-GS} & \multicolumn{2}{c}{SP-GS+NG}  \\
        & & Train$\downarrow$ & FPS$\uparrow$ &Train$\downarrow$ & FPS$\uparrow$  \\ \midrule
        3D Printer  & 151K & 86m  & 149.84 & +15m  & 31.41 \\ 
        Broom       & 565K & 329m & 107.26 & +34m  & 6.6 \\ 
        Chicken     & 153K & 71m  & 146.04 & +11m  & 22.24 \\ 
        Peel Banana & 404K & 180m &  68.30 & +17m  & 10.15 \\ \midrule
        average     & 318K & 167m & 117.86 & +19m  & 17.6 \\ \bottomrule
    \end{tabular}
    }
\end{table}

\subsection{NeRF-DS Dataset.}  \label{sec-ap:NeRF-DS-results}
For the NeRF-DS Dataset, Tab.\ref{tab:NeRF-DS-per-scene} reports the per scene results. 

\subsection{Dynamic Scene Dataset}
We further compare our approach against other methods using the NVIDIA Dynamic Scene Dataset~\cite{Yoon2020NovelVS}, which is composed of 7 video sequences. These sequences are captured with 12 cameras (GoPro Black Edition) utilizing a static camera rig. All cameras concurrently capture images at 12 different time steps. Except for the densify and prune interval, we train our approaches on this dataset using the same configuration as the one employed for the HyperNeRF Dataset. In the initial 15k iterations, we densify 3D Gaussians every 1000 iterations, prune 3D Gaussians every 8000 iterations, and reset opacity every 3000 iterations.

Table \ref{tab:NVIDIA-per-scene} presents the results of quantitative comparisons. In comparison to state-of-the-art methods, our approach also exhibits competitive visual quality.

\section{Additional Experiments}\label{sec-ap:more-exp}
In this section, we conduct additional experiments to investigate key components and factors utilized in our method, aiming to enhance our understand of the mechanism and illustrate its efficacy. %It is important to note that certain modifications have been introduced, resulting in slightly inferior results compared to those presented in the paper.

\subsection{The Loss Weights}
There are three hyperparameters (\ie, $\spP^c$, $\spR^t$, and $\spT^t$) to balance the weights of loss terms. As illustrated in Tab. \ref{tab:loss-weights}, we conduct experiments to showcase the impact of these hyperparameters. It should be emphasized that $\lambda_{\cdot} = 0$ denotes the exclusion of the respective loss term. The results indicate that there is only a minor effect when varying these hyperparameters over a large range ($10^1$ to $0$).

\begin{table}[htb]
    \centering
    \caption{Ablation Study of the loss weights on D-NeRF dataset.}
    \label{tab:loss-weights}
    \setlength{\tabcolsep}{2pt}
    \begin{tabular}{c|*{56}{c}}
        \toprule 
        $\lambda_{\spP^c}$  & $10^0$ & $10^{-1}$  & $10^{-2}$  & $10^{-3}$  & $10^{-4}$ & 0 \\ \midrule
        PSNR$\uparrow$  & 35.498 & 35.435 & 35.521 & 35.338 & 35.667 & 35.350 \\  
        SSIM$\uparrow$ & 0.9808 & 0.9807 & 0.9806 & 0.9810 & 0.9809 & 0.9809 \\
        LPIPS$\downarrow$  & 0.0198 & 0.0200 & 0.0208 & 0.021 & 0.0202 & 0.0142\\ \bottomrule
        \toprule
        $\lambda_{\spT^t}$ & $10^1$ & $10^0$ & $10^{-1}$  & $10^{-2}$  & $10^{-3}$ & 0\\ \midrule
        PSNR$\uparrow$    & 35.542 & 35.483 & 35.379 & 35.509 & 35.552 & 35.551 \\ 
        SSIM$\uparrow$    & 0.9819 & 0.9816 & 0.9813 & 0.9813 & 0.9819 & 0.9817 \\
        LPIPS$\downarrow$ & 0.0128 & 0.0135 & 0.0130 & 0.0129 & 0.0126 &  0.0129\\
        \bottomrule
        \toprule
        $\lambda_{\spR^t}$ & $10^1$ & $10^0$ & $10^{-1}$  & $10^{-2}$  & $10^{-3}$ & 0 \\ \midrule
        PSNR $\uparrow$ & 35.418 & 35.431 & 35.567 & 35.682 & 35.315 & 35.561 \\ 
        SSIM$\uparrow$ & 0.9813 & 0.9813 & 0.9816 & 0.9814 & 0.9813 & 0.9819 \\
        LPIPS$\downarrow$ & 0.0138 & 0.0134 & 0.0131 & 0.0134 & 0.0135 & 0.0133 \\
        \bottomrule
    \end{tabular}
\end{table}

\subsection{The Model Size}
We investigate the impact of the model size of the superpoints deformation network $\mathcal{F}$. We manipulate the network width (\ie, the dimensions of hidden neurons) and the network depth (\ie, the number of hidden layers), presenting results on the D-NeRF dataset in Table \ref{tab:model-size}. Following NeRF ~\cite{NeRF}, when the network depth exceeds 4, we introduce a skip connection between the inputs and the 5th fully-connected layer. With the exception of the configuration with \textit{width}=64 and \textit{depth}=5, which exhibits diminished performance due to the skip concatenation, the experimental results clearly demonstrate that a larger $\mathcal{F}$ leads to a higher visual quality. Since we only need to predict the deformation of superpoints, increasing the model size will results in only a modest rise in computational expense during training. Therefore, employing a larger superpoints deformation network $\mathcal{F}$ is a viable option to enhance the visual quality of dynamic scenes.

\begin{table}[htb]
    \centering
    \caption{Ablation study of the model size of superpoints deformation network $\mathcal{F}$.}
    \label{tab:model-size}
    \begin{tabular}{cc|ccc}
        \toprule
        width & depth & PSNR $\uparrow$ &SSIM$\uparrow$ & LIPIS$\downarrow$ \\
        \midrule
        64 & 1 &  34.7360 & 0.9797 & 0.0152 \\
        64 & 2 &  35.3632 & 0.9814 & 0.0139 \\
        64 & 3 &  35.5986 & 0.9818 & 0.0131 \\
        64 & 4 &  35.3418 & 0.9803 & 0.0142 \\
        64 & 5 &  27.2319 & 0.9491 & 0.0586 \\
        64 & 6 &  35.4901 & 0.9813 & 0.0139 \\
        64 & 7 &  35.7497 & 0.9818 & 0.0130 \\
        64 & 8 &  35.8021 & 0.9823 & 0.0128 \\
        128 & 8 & 36.1375 & 0.9838 & 0.0126 \\
        256 & 8 & 36.4452 & 0.9837 & 0.0123 \\
        \bottomrule
    \end{tabular}
\end{table}

\subsection{Warm-up Train Stage}
To train SP-GS model for a dynamic scene, there is a warm up training stage, \ie we do not train the superpoint deformation network $\mathcal{F}$ and apply deformation to Gaussians in the first 3k iterations. The stage will generate a coarse shape, which is important for initialization of superpoints. The quantity results on D-NeRF dataset  in Table \ref{tab:ab-warm-up} illustrate the the importance of warm up.

\begin{table}[htb]
    \centering
    \caption{Ablation study of the warm-up train stage on D-NeRF dataset.}
    \label{tab:ab-warm-up}
    \begin{tabular}{c*{4}{c}}
    \toprule
    & PSNR & SSIM & LPIPS \\ \midrule
       w/o warm up  & 21.56 & 0.8979 & 0.1445 \\
        w warm-up   & 37.98 & 0.9876 & 0.0164\\
        \bottomrule
    \end{tabular}
\end{table}

\subsection{Inference}
There are two way to rendering images during inference, \ie, using network $\mathcal{F}$ or interpolation. Table \ref{tab:ab-inter} demonstrates that two way have almost same visual quality, but the FPS of using $\mathcal{F}$ is lower than the FPS using interpolation, \ie 168.01 vs. 219.95.

\begin{table}[htb]
    \centering
    \caption{Ablation study for the ways of inference on D-NeRF dataset. \textit{Lego} is included.}
    \label{tab:ab-inter}
    \begin{tabular}{c*{5}{c}}
    \toprule
    & PSNR & SSIM & LPIPS & FPS \\ \midrule
       using $\mathcal{F}$, Eq. \ref{eq:mlp-F} & 36.2281 & 0.9815 & 0.0124 &  168.01 \\
        interp, Eq. \ref{eq:interp}   & 36.2280 & 0.9815 & 0.0124 & 219.95 \\
        \bottomrule
    \end{tabular}
\end{table}

\begin{table*}[t]
    \centering
    \caption{Pre scene performance on the D-NeRF dataset~\cite{D-NeRF}.}
    \label{tab:DNeRF-pre-scene}
    \resizebox{\linewidth}{!}{
    \setlength{\tabcolsep}{2pt}
    \begin{tabular}{c|*{3}{c}|*{3}{c}|*{3}{c}|*{3}{c}}
    \toprule 
    \multirow{2}{*}{Methods} & \multicolumn{3}{c|}{Hell Warrior}  &\multicolumn{3}{c|}{Mutant} &\multicolumn{3}{c|}{Hook} &  \multicolumn{3}{c}{Bouncing Balls} \\
    & PSNR $\uparrow$ & SSIM$\uparrow$ & LPIPS$\downarrow$& PSNR $\uparrow$ & SSIM$\uparrow$ & LPIPS$\downarrow$& PSNR $\uparrow$ & SSIM$\uparrow$ & LPIPS$\downarrow$& PSNR $\uparrow$ & SSIM$\uparrow$ & LPIPS$\downarrow$ \\
    \midrule
    D-NeRF~\cite{D-NeRF} & 24.06 & 0.9440 & 0.0707 & 30.31 & 0.9672 & 0.0392 & 29.02 & 0.9595 & 0.0546 & 38.17 & 0.9891 & 0.0323\\
    TiNeuVox~\cite{TiNeuVox} & 27.10 & 0.9638 & 0.0768 & 31.87 & 0.9607 & 0.0474 & 30.61 & 0.9599 & 0.0592& 40.23& 0.9926& 0.0416 \\
    Tensor4D~\cite{Tensor4D} & 31.26 & 0.9254 & 0.0735 & 29.11 & 0.9451 & 0.0601 & 28.63 & 0.9433 & 0.0636 & 24.47 & 0.9622 & 0.0437 \\
    K-Planes~\cite{KPlanes} & 24.58 & 0.9520 & 0.0824 & 32.50 & 0.9713 & 0.0362 & 28.12 & 0.9489 & 0.0662 & 40.05 & 0.9934 & 0.0322 \\
    HexPlane~\cite{HexPlane} & 24.24 & 0.94 & 0.07 & 33.79 & 0.98 & 0.03 & 28.71 & 0.96 & 0.05 & 39.69 & 0.99 & 0.03 \\
    Ti-DNeRF~\cite{TI-DNeRF} & 25.40 & 0.953 & 0.0682 & 34.70 & 0.983& 0.0226 & 28.76 & 0.960 & 0.0496 & 43.32 & 0.996 & 0.0203 \\
    3D-GS~\cite{3D-GS} & 29.89 & 0.9143 & 0.1113 & 24.50 & 0.9331 & 0.0585 & 21.70 & 0.8864 & 0.1040 & 23.20 & 0.9586 & 0.0608 \\
    D-3D-GS~\cite{Deformable3DGS} & 41.41 & 0.9870 & 0.0115 & 42.61 & 0.9950 & 0.0020 & 37.09 & 0.9858 & 0.0079 & 40.95 & 0.9953 & 0.0027  \\
    4D-GS~\cite{RealTime4DGS} & 28.77 & 0.9729 & 0.0241 & 37.43 & 0.9874 & 0.0092 & 33.01 & 0.9763 & 0.0163 & 40.78 & 0.9942 & 0.0060  \\
    % SC-GS\cite{SC-GS} & 42.93 & 0.994 & 0.0155 & 45.19 & 0.999 & 0.0028 & 39.87 & 0.997 & 0.0076 & 44.91 & 0.998 & 0.0166 \\
    \textbf{SP-GS (ours)}   & 40.19 & 0.9894 & 0.0066 & 39.43 & 0.9868 & 0.0164 & 35.36 & 0.9804 & 0.0187 & 40.53 & 0.9831 & 0.0326 \\
    \textbf{SP-GS+NG(ours)} & 39.01 & 0.9938 & 0.0043 & 41.02 & 0.9890 & 0.0112 & 35.35 & 0.9827 & 0.0138 & 41.65 & 0.9762 & 0.0278  \\
    \bottomrule
    \multirow{2}{*}{Methods} & \multicolumn{3}{c|}{Lego}  &\multicolumn{3}{c|}{T-Rex} &\multicolumn{3}{c|}{Stand Up} &  \multicolumn{3}{c}{Jumping Jacks} \\
    & PSNR $\uparrow$ & SSIM$\uparrow$ & LPIPS$\downarrow$& PSNR $\uparrow$ & SSIM$\uparrow$ & LPIPS$\downarrow$& PSNR $\uparrow$ & SSIM$\uparrow$ & LPIPS$\downarrow$& PSNR $\uparrow$ & SSIM$\uparrow$ & LPIPS$\downarrow$ \\
    \midrule 
    D-NeRF~\cite{D-NeRF} &   25.56&  0.9363 & 0.0821 & 30.61 & 0.9671 & 0.0535 & 33.13 & 0.9781 & 0.0355 & 32.70 & 0.9779 & 0.0388  \\
    TiNeuVox~\cite{TiNeuVox} & 26.64 & 0.9258 & 0.0877 & 31.25 & 0.9666 & 0.0478 & 34.61 & 0.9797 & 0.0326 & 33.49 & 0.9771 & 0.0408  \\
    Tensor4D~\cite{Tensor4D} & 23.24 & 0.9183 & 0.0721 & 23.86&  0.9351 & 0.0544 & 30.56 & 0.9581 & 0.0363 & 24.20 & 0.9253 & 0.0667 \\
    K-Planes~\cite{KPlanes} & 28.91 & 0.9695 & 0.0331 & 30.43 & 0.9737 & 0.0343 & 33.10 & 0.9793 & 0.0310 & 31.11 & 0.9708 & 0.0468 \\
    TI-DNeRF~\cite{TI-DNeRF} &25.33 & 0.943 & 0.0413 & 33.06 & 0.982 & 0.0212 & 36.27 & 0.988 & 0.0159 &35.03& 0.985 & 0.0249 \\
    HexPlane~\cite{HexPlane} &
    25.22 & 0.94 & 0.04 & 30.67 & 0.98 & 0.03 & 34.36 & 0.98 & 0.02 & 31.65 & 0.97 & 0.04  \\
    3D-GS~\cite{3D-GS} & 23.04 & 0.9288 & 0.0521 & 21.91 & 0.9536 & 0.0552 & 21.91 & 0.9299 & 0.0893 & 20.64 & 0.9292 & 0.1065 \\
    D-3D-GS~\cite{Deformable3DGS} & 24.91 & 0.9426 & 0.0299 & 37.67 & 0.9929 & 0.0041 & 44.30 & 0.9947 & 0.0031 & 37.59 & 0.9893 & 0.0085  \\
    4D-GS~\cite{RealTime4DGS} &  25.04 & 0.9362 & 0.0382 & 33.61 & 0.9828 & 0.0136 & 38.11 & 0.9896 & 0.0072 & 35.44 & 0.9853 & 0.0127\\
    % SC-GS\cite{3D-GS} & - & - & - & 41.24 & 0.998 & 0.0046 & 47.89 & 0.999 & 0.0023 & 41.13 & 0.998 & 0.0067 \\
    \textbf{SP-GS (ours)}    & 24.48 & 0.9390 & 0.0331 & 32.69 & 0.9861 & 0.0243 & 42.07 & 0.9926 & 0.0096 & 35.56 & 0.9950 & 0.0069 \\
    \textbf{SP-GS+NG (ours)} & 28.58 & 0.9518 & 0.0331 & 34.47 & 0.9839 & 0.0182 & 42.12 & 0.9925 & 0.0065 & 34.32 & 0.9959 & 0.0064  \\
    \bottomrule
    \end{tabular}
    }
\end{table*}

\begin{table*}[th]
    \centering
    \caption{Per scene quantitative comparisons on the HyperNeRF~\cite{HyperNeRF} dataset.}
    \label{tab:Hyper-pre-scene}
    \resizebox{\linewidth}{!}{
    \setlength{\tabcolsep}{2pt}
    \begin{tabular}{c|*{3}{c}|*{3}{c}|*{3}{c}|*{3}{c}|*{3}{c}}
        \toprule
        \multirow{2}{*}{Methods} &  \multicolumn{3}{c|}{Broom} & \multicolumn{3}{c|}{3D Printer} & \multicolumn{3}{c|}{Chicken}  & \multicolumn{3}{c|}{Peel banana}& \multicolumn{3}{c}{Mean} \\
         & \small PSNR $\uparrow$ &  \small MS-SSIM$\uparrow$ &  \small LIPIS$\downarrow$ & \small PSNR $\uparrow$ &  \small MS-SSIM$\uparrow$ & \small LIPIS$\downarrow$ & \small PSNR $\uparrow$ & \small MS-SSIM$\uparrow$ & \small LIPIS$\downarrow$  & \small PSNR $\uparrow$ & \small MS-SSIM$\uparrow$ & \small LIPIS$\downarrow$ & \small PSNR $\uparrow$ & \small MS-SSIM$\uparrow$ & \small LIPIS$\downarrow$  \\ \midrule

        NeRF~\cite{NeRF} & 19.9 & 0.653 & 0.692 & 20.7 & 0.780 & 0.357 & 19.9 & 0.777 & 0.325 & 20.0 & 0.739 & 0.413 & 20.1 & 0.735  & 0.424 \\
        NV~\cite{NV} & 17.7 &  0.623 & 0.360 & 16.2 & 0.665 & 0.330 & 17.6 & 0.615 & 0.336 & 15.9 & 0.380 & 0.413 & 16.9 & 0.571 & -  \\
        NSFF~\cite{NSFF}  & 26.1 & 0.871 & 0.284 & 27.7 & 0.947 & 0.125 & 26.9 & 0.944 & 0.106 & 24.6 & 0.902 & 0.198 & 26.3 & 0.916 & -  \\
        Nerfies~\cite{Nerfies}  & 19.2 & 0.567 & 0.325 & 20.6 & 0.830 & 0.108 & 26.7 & 0.943 & 0.0777 & 22.4 & 0.872 & 0.147  & 22.2 & 0.803 & - \\
        HyperNeRF~\cite{HyperNeRF}  & 19.3 & 0.591 & 0.296 & 20.0 & 0.821 & 0.111 & 26.9 & 0.948 & 0.0787 & 23.3 & 0.896 & 0.133 & 22.4 & 0.814  & - \\
        TiNeuVox-S~\cite{TiNeuVox}  & 21.9 & 0.707 & -& 22.7 & 0.836 & -& 27.0 & 0.929 & - & 22.1&0.780&-  & 23.4 & 0.813 &- \\
        TiNeuVox-B~\cite{TiNeuVox} & 21.5 & 0.686 &- & 22.8 & 0.841 & -&  28.3 & 0.947 & - & 24.4&0.873& - & 24.3 & 0.837 & - \\ 
        NDVG~\cite{NDVG} & 22.4 & 0.839 & - & 21.5 & 0.703 & - & 27.1  & 0.939 & - & 22.8 & 0.828 & - & 23.3 & 0.823 & - \\
        TI-DNeRF~\cite{TI-DNeRF} & 20.48 & 0.685 & - & 20.38 & 0.678 & - &21.89&0.869 & - & 28.87 & 0.965 &-   & 24.35 & 0.866 &- \\
        NeRFPlayer~\cite{NeRFPlayer}  & 21.7 & 0.635 & - &22.9&0.810& -  & 26.3 & 0.905 & - & 24.0 & 0.863 & - & 23.7 & 0.803 & -\\
        3D-GS~\cite{3D-GS} & 19.74 & 0.4949 & 0.3745 & 19.26 & 0.6686  & 0.4281  & 22.51 & 0.7954 & 0.3307 & 19.54 & 0.6688 & 0.2339 & 20.26 & 0.6569  & 0.3418 \\
        4D-GS~\cite{RealTime4DGS} & 22.01 & 0.6883 & 0.5448 & 21.98 & 0.8038 & 0.2763 & 27.58 & 0.9333 & 0.1468 & 28.52 & 0.9254 & 0.198 & 25.02 & 0.8377 & 0.2915  \\
        \textbf{SP-GS (ours)}    & 20.07 & 0.6004 & 0.3430 & 24.31 & 0.8719 & 0.2312 & 30.81 & 0.9550 & 0.1262 & 27.23 & 0.9341 & 0.1286 & 25.61 & 0.8404 & 0.2073 \\
        \textbf{SP-GS+NR (ours)} & 22.76 & 0.7794 & 0.2812 & 24.88 & 0.8836	& 0.2100 & 31.47 & 0.9609 & 0.1122 & 28.01 & 0.9442 & 0.1186 & 26.78 & 0.8920 & 0.1805 \\
        \bottomrule
    \end{tabular}
    }
\end{table*}

\begin{table*}
    \centering
    \caption{Quantitative comparison on NeRF-DS dataset~\cite{NeRF-DS} pre-scene. LPIPS use the VGG network.}
    \label{tab:NeRF-DS-per-scene}
    \setlength{\tabcolsep}{2pt}
    \resizebox{\linewidth}{!}{
    \begin{tabular}{c|*{3}{c}|*{3}{c}|*{3}{c}|*{3}{c}}
        \toprule
        \multirow{2}{*}{Method} & \multicolumn{3}{c|}{Sieve} & \multicolumn{3}{c|}{Plate} & \multicolumn{3}{c|}{Bell} & \multicolumn{3}{c}{Press} \\
         & PSNR$\uparrow$ & MS-SSIM$\uparrow$ & LPIPS$\downarrow$ & PSNR$\uparrow$ & MS-SSIM$\uparrow$ & LPIPS$\downarrow$ & PSNR$\uparrow$ & MS-SSIM$\uparrow$ & LPIPS$\downarrow$ & PSNR$\uparrow$ & MS-SSIM$\uparrow$ & LPIPS$\downarrow$ \\
        \midrule
TiNeuVox  & 21.49 & 0.8265 & 0.3176 & 20.58 & 0.8027 & 0.3317 & 23.08 & 0.8242 & 0.2568 & 24.47 & 0.8613 & 0.3001 \\
HyperNeRF & 25.43 & 0.8798 & 0.1645 & 18.93 & 0.7709 & 0.2940 & 23.06 & 0.8097 & 0.2052 & 26.15 & 0.8897 & 0.1959 \\
NeRF-DS   & 25.78 & 0.8900 & 0.1472 & 20.54 & 0.8042 & 0.1996 & 23.19 & 0.8212 & 0.1867 & 25.72 & 0.8618 & 0.2047 \\
3D-GS    & 23.16 & 0.8203 & 0.2247 & 16.14 & 0.6970 & 0.4093 & 21.01 & 0.7885 & 0.2503 & 22.89 & 0.8163 & 0.2904\\
D-3D-GS   & 25.01 & 0.867 & 0.1509 & 20.16 & 0.8037 & 0.2243 & 25.38 & 0.8469 & 0.1551 & 25.59 & 0.8601 & 0.1955 \\
\textbf{SP-GS(ours)}    & 25.62 & 0.8651 & 0.1631 & 18.91 & 0.7725 & 0.2767 & 25.20 & 0.8430 & 0.1704 & 24.34 & 0.846 & 0.2157\\
\textbf{SP-GS+NG(ours)} & 25.39 & 0.8599 & 0.1667 & 19.81 & 0.7849 & 0.2538 & 24.97 & 0.8421 & 0.1782 & 24.93 & 0.861 & 0.2073 \\
% SC-GS \cite{SC-GS} & 26.0 & 0.919 & 0.114 & 20.2 & 0.837 & 0.202 & 25.1 & 0.918 & 0.117 & 26.6 & 0.901 & 0.135 \\
        \toprule
        \multirow{2}{*}{Method} & \multicolumn{3}{c|}{Cup} & \multicolumn{3}{c|}{As} & \multicolumn{3}{c|}{Basin} & \multicolumn{3}{c}{Mean} \\
         & PSNR$\uparrow$ & MS-SSIM$\uparrow$ & LPIPS$\downarrow$ & PSNR$\uparrow$ & MS-SSIM$\uparrow$ & LPIPS$\downarrow$ & PSNR$\uparrow$ & MS-SSIM$\uparrow$ & LPIPS$\downarrow$ & PSNR$\uparrow$ & MS-SSIM$\uparrow$ & LPIPS$\downarrow$ \\
        \midrule
TiNeuVox  & 19.71 & 0.8109 & 0.3643 & 21.26 & 0.8289 & 0.3967 & 20.66 & 0.8145 & 0.2690 & 21.61 & 0.8234 & 0.2766 \\
HyperNeRF & 24.59 & 0.8770 & 0.1650 & 25.58 & 0.8949 & 0.1777 & 20.41 & 0.8199 & 0.1911 & 23.45 & 0.8488 & 0.1990 \\
NeRF-DS   & 24.91 & 0.8741 & 0.1737 & 25.13 & 0.8778 & 0.1741 & 19.96 & 0.8166 & 0.1855 & 23.60 & 0.8494 & 0.1816 \\
3D-GS     & 21.71 & 0.8304 & 0.2548 & 22.69 & 0.8017 & 0.2994 & 18.42 & 0.7170 & 0.3153 & 20.29 & 0.7816 & 0.2920 \\
D-3D-GS   & 24.54 & 0.8848 & 0.1583 & 26.15 & 0.8816 & 0.1829 & 19.61 & 0.7879 & 0.1897 & 23.78 & 0.8474 & 0.1795  \\
\textbf{SP-GS(ours)}   & 24.43 & 0.8823 & 0.1728 & 24.44 & 0.8626 & 0.2255 & 19.09 & 0.7627 & 0.2189 & 23.15 & 0.8335 & 0.2062 \\
\textbf{SP-GS+NG(ours)}& 23.66 & 0.8738 & 0.1853 & 25.16 & 0.8650 & 0.2246 & 19.36 & 0.7667 & 0.2429 & 23.33 & 0.8362 & 0.2084 
\\
% SC-GS\cite{SC-GS} & 24.5 & 0.916 & 0.115 & 26.2 & 0.898 & 0.142 & 19.6 & 0.846 & 0.154 & 24.1 & 0.891 & 0.140 \\
        \bottomrule
    \end{tabular}
    }
\end{table*}

\begin{table*}
    \centering
    \caption{Quantitative results on NVIDIA Dynamic Scene dataset~\cite{Yoon2020NovelVS}.}
    \label{tab:NVIDIA-per-scene}
    \resizebox{\linewidth}{!}{
    \setlength{\tabcolsep}{2pt}
    \begin{tabular}{c|*{3}{c}|*{3}{c}|*{3}{c}|*{3}{c}}
        \toprule
        \multirow{2}{*}{Methods} &  \multicolumn{3}{c|}{Jumping} & \multicolumn{3}{c|}{Skating} & \multicolumn{3}{c|}{Truck} & \multicolumn{3}{c}{Umbrella}   \\ 
        & \small PSNR$\uparrow$ & \small SSIM$\uparrow$ & \small LPIPS$\downarrow$  
        & \small PSNR$\uparrow$ & \small SSIM$\uparrow$ & \small LPIPS$\downarrow$  
        & \small PSNR$\uparrow$ & \small SSIM$\uparrow$ & \small LPIPS$\downarrow$  
        & \small PSNR$\uparrow$ & \small SSIM$\uparrow$ & \small LPIPS$\downarrow$  
        \\
        \midrule
    NeRF~\cite{NeRF}+time &  16.72 & 0.42 & 0.489 & 19.23 & 0.46 & 0.542 & 17.17 & 0.39 & 0.403 & 17.17 & - & 0.752  \\
    
    D-NeRF~\cite{D-NeRF} & 21.0 & 0.68 & 0.21 & 20.8 & 0.62 & 0.35 & 22.9 & 0.71 & 0.15   &  -&-&-  \\
    NR-NeRF~\cite{NR-NeRF} & 19.38 & 0.61 & 0.295 & 23.29 & 0.72 & 0.234 & 19.02 & 0.44 & 0.453 & 19.26 & - &0.427   \\
    HyperNeRF~\cite{HyperNeRF} &  17.1 & 0.45 & 0.32 & 20.6 & 0.58 & 0.19  &19.4 & 0.43 & 0.21  &  -&-&- \\
    TiNeuVox~\cite{TiNeuVox} &  19.7 & 0.60 & 0.26 & 21.9 & 0.68 & 0.16 & 22.9 & 0.63 & 0.19   &  -&-&- \\
    NSFF~\cite{NSFF}  &  24.12 & 0.80 & 0.144 & 28.90 & 0.88 & 0.124 & 25.94 & 0.76 & 0.171 & 22.58 & - & 0.302  \\
    DVS~\cite{DVS} &  23.23 & 0.83 & 0.144 & 28.90 & 0.94 & 0.124 & 25.78 & 0.86 & 0.134 & 23.15 & - & 0.146  \\
    RoDynRF~\cite{RoDynRF} &  25.66 & 0.84 & 0.071 & 28.68 & 0.93 & 0.040 & 29.13 & 0.89 & 0.063 & 24.26 & - & 0.063  \\
    Point-DynRF~\cite{Point-DynRF} & 23.6 & 0.90 & 0.14 & 29.6 & 0.96 & 0.04 & 28.5 & 0.94 & 0.08  & -&-&-  \\
    \textbf{SP-GS (ours)}   & 22.13 &0.7484&0.4675 & 29.21 & 0.9079 & 0.2360 & 27.38 &0.8401&0.1898 & 24.88 & 0.6568&0.3231 \\
    \textbf{SP-GS+NG(ours)} & 23.41 &0.8104&0.3267 & 29.54 & 0.9124 & 0.2323 & 27.62 &0.8440&0.1860 & 25.18 & 0.6617&0.3200 \\
     \bottomrule
    \toprule
    \multirow{2}{*}{Methods} 
    &\multicolumn{3}{c|}{ Balloon1} & \multicolumn{3}{c|}{Balloon2} & \multicolumn{3}{c|}{Playground} & \multicolumn{3}{c}{Avg} \\ 
    & \small PSNR$\uparrow$ & \small SSIM$\uparrow$ & \small LPIPS$\downarrow$  
    & \small PSNR$\uparrow$ & \small SSIM$\uparrow$ & \small LPIPS$\downarrow$  
    & \small PSNR$\uparrow$ & \small SSIM$\uparrow$ & \small LPIPS$\downarrow$  
    & \small PSNR$\uparrow$ & \small SSIM$\uparrow$ & \small LPIPS$\downarrow$   \\
    \midrule
NeRF~\cite{NeRF}+time & 17.33 & 0.40 & 0.304    & 19.67 & 0.54 & 0.236 & 13.80 & 0.18 & 0.444 & 17.30 & 0.40 & 0.453 \\
D-NeRF~\cite{D-NeRF} & 18.0 & 0.44 & 0.28 & 19.8 & 0.52 & 0.30 & 19.4 & 0.65 & 0.17 & 20.4 & 0.59 & 0.24 \\
NR-NeRF~\cite{NR-NeRF} & 16.98 & 0.34 & 0.225  & 22.23 & 0.70 & 0.212 & 14.24 & 0.19 & 0.336 & 19.20 & 0.50 & 0.330 \\
HyperNeRF~\cite{HyperNeRF} &12.8 & 0.13 & 0.56  & 15.4 & 0.20 & 0.44 & 12.3 & 0.11 & 0.52  &16.3 & 0.32 & 0.37 \\
TiNeuVox~\cite{TiNeuVox}& 16.2 & 0.34 & 0.37  & 18.1 & 0.41 & 0.29  & 12.6 & 0.14 & 0.46  & 18.6 & 0.47 & 0.29\\
NSFF~\cite{NSFF}  & 21.40 & 0.69 & 0.225 & 24.09 & 0.73 & 0.228 & 20.91 & 0.70 & 0.220 & 23.99 & 0.76 & 0.205 \\
DVS~\cite{DVS}  & 21.47 & 0.75 & 0.125 & 25.97 & 0.85 & 0.059 & 23.65 & 0.85 & 0.093 & 24.74 & 0.85 & 0.118 \\
RoDynRF~\cite{RoDynRF}& 22.37 & 0.76 & 0.103  & 26.19 & 0.84 & 0.054 & 24.96 & 0.89 & 0.048 & 25.89 & 0.86 & 0.065 \\
Point-DynRF~\cite{Point-DynRF}& 21.7 & 0.88 & 0.12 & 26.2 & 0.92 & 0.06 & 22.2 & 0.91 & 0.09 & 25.3 & 0.92 & 0.08 \\ 
\textbf{SP-GS (ours)}   & 24.36 & 0.8783 & 0.1802 & 29.65 &0.9059&0.0965 & 22.29 &0.7721&0.2338 & 25.70 & 0.8156 & 0.2467\\
\textbf{SP-GS+NG(ours)} & 24.96 & 0.8811 & 0.1808 & 26.31 &0.7882&0.2291 & 20.28 &0.7453&0.3488 & 25.33 & 0.8062 & 0.2605\\
\bottomrule
    \end{tabular}
    }
\end{table*}

% \cleardoublepage